\documentclass{article}

\usepackage{iclr2027_conference,times}
\iclrfinalcopy  

\usepackage{amsmath,amssymb,amsfonts,mathtools,bm}
\usepackage{booktabs,multirow,array,tabularx,makecell}
\usepackage{graphicx,adjustbox,subcaption}
\usepackage{xcolor,colortbl}
\usepackage{enumitem}
\usepackage{microtype}
\usepackage[hidelinks]{hyperref}
\hypersetup{hidelinks}
\usepackage{url}
\usepackage{tikz}
\usepackage{wrapfig}
\usepackage[most]{tcolorbox}
\newtcolorbox{promptbox}[1][]{
    enhanced,
    breakable,
    colback=gray!4,
    colframe=gray!55,
    boxrule=0.6pt,
    arc=2pt,
    left=7pt,
    right=7pt,
    top=6pt,
    bottom=6pt,
    fontupper=\small,
    fonttitle=\small\bfseries,
    title=#1,
    before skip=8pt,
    after skip=8pt,
}
\usetikzlibrary{arrows.meta,positioning,fit,calc}

\definecolor{placeholderbg}{RGB}{250,246,232}
\definecolor{placeholderborder}{RGB}{176,128,0}
\definecolor{phaseblue}{RGB}{231,241,250}
\definecolor{phasegreen}{RGB}{234,246,237}
\definecolor{phasegray}{RGB}{244,244,244}

\newcommand{\benchmark}{\textsc{Common-TSQA}}

\newcommand{\Original}{Original}

\newcommand{\Flat}{Flat}

\newcolumntype{Y}{>{\raggedright\arraybackslash}X}

\newcommand{\tomni}{TimeOmni-1}
\newcommand{\chatts}{ChatTS}
\newcommand{\tovl}{TimeOmni-VL}
\newcommand{\timemqa}{Time-MQA}
\newcommand{\tsr}{TSR-Suite}

\title{Do Time-Series QA Systems Read the Time Series? Evidence Use and Reasoning Reliability}

\author{Zhuomin Chen\textsuperscript{1},
        Jingchao Ni\textsuperscript{2},
        Xu Zheng\textsuperscript{1},
        Janki Bhimani\textsuperscript{1}, 
        Mo Sha\textsuperscript{1}, 
        Wei Cheng\textsuperscript{3}, \\
\textbf{Dongsheng Luo}\textsuperscript{4} \\
\textsuperscript{1}Florida International University\\
\textsuperscript{2}University of Houston\\
\textsuperscript{3}NEC Laboratories America\\
\textsuperscript{4}Singapore Management University\\
\texttt{\{zchen051,xzhen019,jbhimani,msha\}@fiu.edu}\\
\texttt{jni7@uh.edu}\\
\texttt{weicheng@nec-labs.com}\\
\texttt{luodongsheng01@gmail.com}
}

\begin{document}
\maketitle

\begin{abstract}
In recent years, time-series question answering (QA) systems have made significant progress. However, generating a correct answer does not show whether retaining the supplied numerical series improves task performance, nor whether the prediction is sensitive to changes in that input.  While some systems provide rationales, answer accuracy also does not show whether their numerical claims are grounded in the supplied series or whether the stated inference is valid.
In this work, we focus on evaluating four time-series QA systems: \tomni{}, \chatts{}, \tovl{}, and \timemqa{}. 
First, for three systems with released evaluation data, we reproduce their reported results and compare the performance of the systems with their backbones.
Then, we introduce a benchmark named \benchmark{}, which collects public evaluation datasets from existing time-series benchmarks and unifies their sample representation, task definitions, and answer schemas, while evaluating each system through its own interface under common evaluation criteria.
The evaluation uses the original condition and six interventions while keeping the question and target fixed.
Our analysis shows that aggregate performance alone can obscure how systems use numerical evidence. Similar task-level scores can arise despite substantial changes in individual predictions. Some interventions induce simple fallback behavior rather than preserved task ability. 
We also evaluate rationales for factual grounding, inference validity, and consistency with the final answer.
We find that rationales often contain time-series claims unsupported by the input. Moreover, the rationale audit shows that agreement between a rationale and its final answer can coexist with incorrect numerical descriptions or invalid intermediate inferences.
\end{abstract}

\section{Introduction}
\label{sec:intro}

\begin{figure}[t]
\centering
\includegraphics[width=\linewidth]{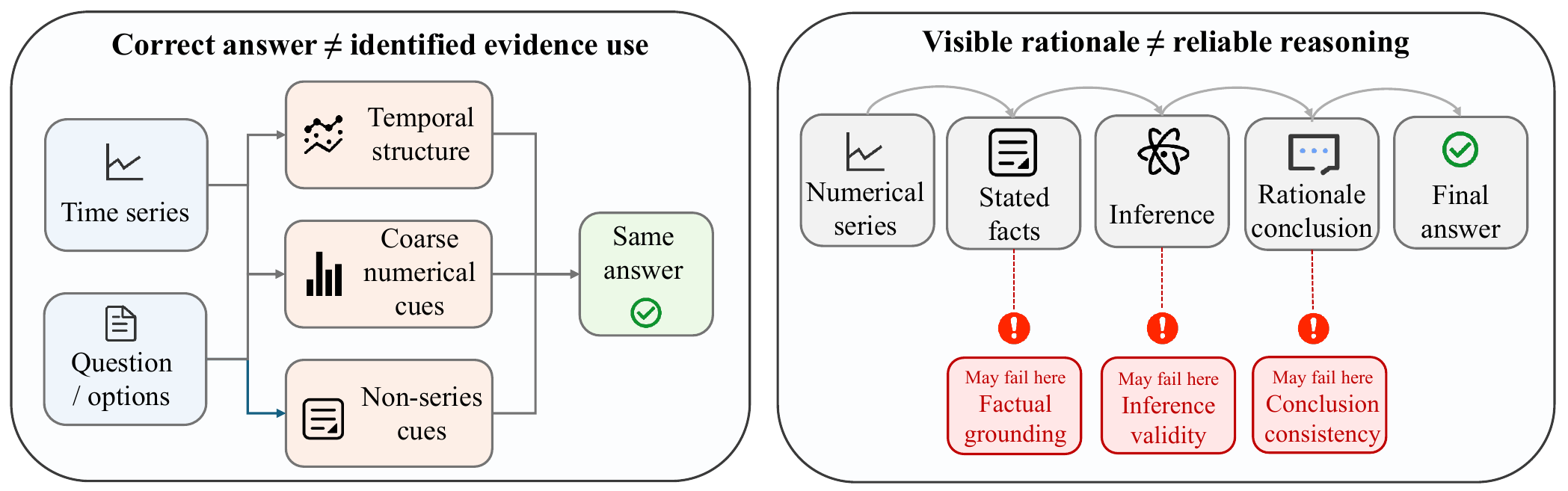}
\caption{Final-answer correctness and rationales leave distinct ambiguities. (a) The same answer can be compatible with temporal structure, coarse numerical cues, or non-series cues. (b) Even when a rationale is visible, failures can arise in factual grounding, inference validity, or consistency with the final answer.}
\label{fig:intro}
\end{figure}

Recent time-series question-answering (QA) systems extend large language
models to tasks such as forecasting, pattern recognition, causal diagnosis,
cross-series comparison, and decision-making from numerical time series
\citep{xie2025chatts,kong2025timemqa,guan2026timeomni1,guan2026timeomnivl}.
Specialized systems report substantial improvements over their underlying
language-model backbones, but final-answer accuracy alone does not identify
what evidence supported a prediction. As illustrated in Figure~\ref{fig:intro}, the same benchmark answer may be compatible with temporal structure in the series, coarse numerical cues, or information already available from the question, answer choices, and surrounding context. A visible natural-language rationale does not resolve this ambiguity: it may appear coherent while misdescribing the supplied series or drawing an invalid conclusion from its own stated premises.


Prior work has questioned whether strong time-series performance  reflects meaningful use of the numerical input, and separate work has examined the correctness of generated time-series explanations
\citep{merrill2024language,sivalingam2026tsquerybench}. These observations motivate a question: when the numerical evidence for the same item is changed, how does the prediction respond, and does the response remain appropriate for the modified evidence? For systems that expose visible rationales, a question is whether the numerical facts stated in the reasoning are supported by the supplied series, and whether the subsequent conclusion follows from those facts.

We study these questions on four time-series question answering (QA) systems: \tomni{}~\citep{guan2026timeomni1}, \chatts{}~\citep{xie2025chatts}, \tovl{}~\citep{guan2026timeomnivl}, and \timemqa{}~\citep{kong2025timemqa}. 
For the three systems for which their original evaluation data are available, we first reproduce the results and compare them to their corresponding backbones. 
We then construct \benchmark{}, a common evaluation benchmark assembled
from public time-series evaluation datasets. Rather than introducing new task semantics, \benchmark{} retains the source-defined questions, targets, and task objectives while standardizing the representation and scoring needed for cross-system evaluation. Each system is evaluated through its native input and output interface, and outputs are scored using the same task-specific criteria.
Each sample is evaluated under original condition and six interventions with fixed question and original target.

Our evaluation separates three questions that are often conflated by
final-answer performance. First, we ask whether specialization improves the systems' native task performance (RQ1). Second, we examine how predictions respond when the numerical evidence is changed, including whether apparent performance changes reflect prediction instability, simple fallback behavior, or an answer that should itself change under the intervention (RQ2). Third, for systems with visible reasoning, we audit the path from numerical evidence to stated facts, inference, and final conclusion (RQ3). 



Our main contributions are:
\begin{itemize}
    \item We construct \benchmark{}, a common evaluation benchmark across six time-series QA tasks, allowing four specialized systems to be evaluated on the same samples and scoring criteria.

    \item We develop an intervention-based analysis of numerical-evidence use. The analysis distinguishes prediction turnover, fallback behavior under evidence removal or simplification, and cases in which the intervention changes the answer implied by the numerical input.

    \item We audit visible rationales for factual grounding, inference validity, and consistency with the final answer, identifying failures at distinct stages of the evidence-to-answer reasoning path.
    
\end{itemize}

\section{Related Work}
\label{sec:related}

\paragraph{Time-series language models.}
Recent time-series language models extend beyond forecasting to a broad range of understanding and reasoning tasks. \textsc{ChatTS} \citep{xie2025chatts} uses a dedicated time-series encoder and aligns its temporal representations with a language model through synthetic alignment data and supervised fine-tuning. \textsc{ChatTime} \citep{wang2025chattime} treats time series as a ``foreign language'' and models numerical and textual data in a unified framework. \textsc{Time-MQA} \citep{kong2025timemqa} builds a large-scale multi-task corpus covering forecasting, imputation, anomaly detection, classification, and open-ended question answering, while \textsc{TimeOmni-1} \citep{guan2026timeomni1} combines supervised fine-tuning and reinforcement learning for scenario understanding, causality discovery, event-aware forecasting, and decision-making. 
More recent systems further develop task-specific mechanisms for time-series reasoning. \textsc{PATRA} \citep{lu2026patra} introduces pattern-aware alignment for trend and seasonality together with a task-aware balanced reward for heterogeneous TSQA tasks. \textsc{ARTIST} \citep{messica2026artist} interleaves reasoning with question-adaptive temporal segment selection through a controller--reasoner architecture. \textsc{TimeOmni-VL} \citep{guan2026timeomnivl} instead adopts a vision-centric formulation that maps between time series and images for both understanding and numerical generation.

\paragraph{Evidence dependence and behavioral evaluation.}
Prior work has questioned whether strong time-series performance
necessarily reflects effective use of all available inputs. \citet{tan2024useful} ablate the language-model components of OneFitsAll, Time-LLM,
and CALF, finding that removing the pretrained language model or
replacing it with substantially simpler modules often leaves forecasting
performance largely unchanged. \textsc{CiK} \citep{williams2025context} approaches
the question from the data side by constructing forecasting tasks in which
textual context is necessary for accurate prediction. In multimodal
forecasting, \citet{zhang2026when} find that textual information
is most useful when it provides information that cannot be recovered from
the time series alone.
\citet{merrill2024language} study the role of the numerical input more directly.
On an initial multiple-choice benchmark with a single time series, models
retain much of their performance when the series is removed. Then they introduce a harder two-series setting, where access to the series provides only limited benefit for most evaluated models. These studies show that strong task performance does not by itself establish substantial dependence on every available input source.

\paragraph{Faithfulness and reliability of generated reasoning.}
A correct final answer does not guarantee that a rationale provides
a reliable account of the evidence or reasoning leading to that answer.
\citet{jacovi2020faithfulness} distinguish faithfulness from plausibility,
with faithfulness concerning whether an explanation reflects factors that
actually influence a model's prediction. \textsc{ERASER} \citep{deyoung2020eraser}
evaluates rationales through human-identified evidence and its effect on
model predictions. For chain-of-thought reasoning, \citet{lanham2023measuring}
intervene on intermediate reasoning through perturbation and paraphrasing
and measure the resulting changes in the answer. \citet{turpin2023language}
show that models can produce convincing rationales that fail to disclose
the influence of biasing features.
More recent work develops benchmarks for evaluating such failures.
\textsc{FaithCoT-Bench} \citep{faithcotbench2025} studies instance-level faithfulness of chain-of-thought reasoning, while \textsc{RFEval} \citep{rfeval2026} evaluates
reasoning faithfulness under counterfactual interventions. In the
time-series domain, \textsc{TSQueryBench} \citep{sivalingam2026tsquerybench} studies reference-free evaluation of numerical correctness in generated
time-series explanations.

\section{\benchmark{}: A Common Evaluation Benchmark}
\label{sec:benchmark}

Reported results from current time-series reasoning systems are difficult to compare fairly. Some systems do not release the evaluation data used in their papers, while other systems are evaluated on different tasks, datasets, or splits. Therefore, performance differences may depend on the test set rather than the systems. 
We need a shared test set where all systems are evaluated on the same canonical samples, task definitions, targets, and evaluation criteria.
For this purpose, we build \benchmark{}. Instead of creating a new task corpus, we construct it from evaluation sets of existing time-series benchmarks, retaining the original tasks and targets defined by the sources. Using several benchmarks also broadens the coverage of tasks and domains and reduces dependence on the annotation conventions or question formats of any one benchmark. \benchmark{} contains samples from six tasks and is constructed to avoid overlap with the training datasets of the evaluated systems. All systems use the same canonical samples and tasks, with the input formatted according to the interface required by each system. 

The public benchmark sources include GIFT-Eval, SciTS, TIME, 
TSB-AD-U, TSAQA, CaTS-Bench, FactoryBench, SenTSR-Bench, TSRBench, and TSR-Suite-OOD \citep{aksu2025gifteval,wu2026scits,qiao2026time,liu2024elephant,factorybench2026}. Since these samples already have source-defined targets and task semantics, we only standardize what is needed for cross-system evaluation.

For numerical tasks, we keep the source observations and source-defined targets when they can be mapped deterministically to the benchmark schema. For question answering (QA) tasks, we preserve the question meaning, option semantics, and target label. Each record contains the time series, question, options, target, source split, task and domain metadata, and source grouping information.
Details on the sources, conversion rules, and task-domain composition can be found in Appendix~\ref{app:benchmark}.

\section{Evaluation Settings}
\label{sec:evaluation-setting}

\subsection{Systems}

We evaluate four time-series QA systems: \tomni{}~\citep{guan2026timeomni1}, \chatts{}~\citep{xie2025chatts}, \tovl{}~\citep{guan2026timeomnivl}, and \timemqa{}~\citep{kong2025timemqa}.
Each checkpoint is evaluated through its native input interface. Thus, the systems receive the same canonical sample and task content, while the numerical series and accompanying information are rendered in the format required by each system. Details on the backbones, training procedures, and native reproduction are provided in Appendix~\ref{app:native}.

\subsection{Evidence interventions}
\label{sec:interventions}

\begin{figure}[t]
\centering
\includegraphics[width=\linewidth]{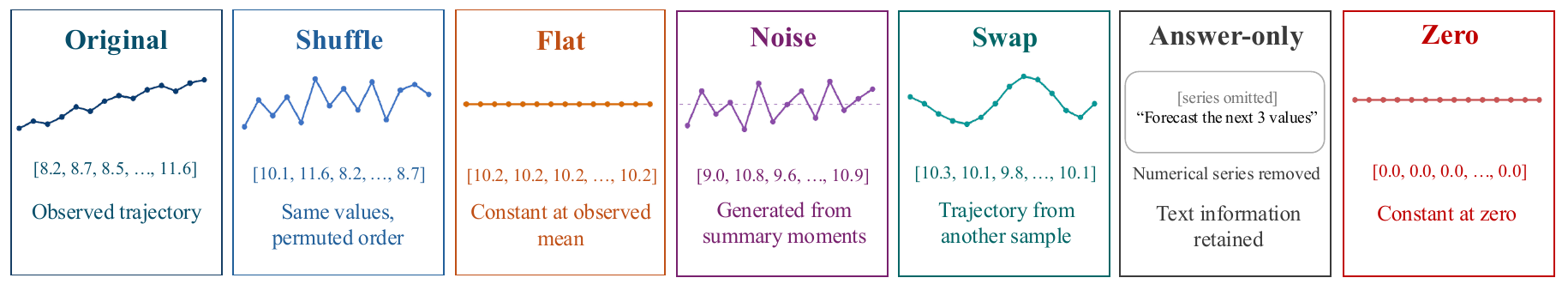}
\caption{The original input and six interventions to the numerical evidence.
Non-series input content is fixed across conditions.
}
\label{fig:intervention}
\end{figure}

For each sample, we evaluate the system under the original condition and six interventions to the numerical series. The interventions modify only the numerical evidence; the question, answer choices and their order, task context, ground-truth label, and other non-series information are held fixed.
Figure~\ref{fig:intervention} illustrates the seven conditions.


The six interventions are:
\begin{itemize}
    \item \textbf{Shuffle} independently permutes the observed values
    within each series, preserving its length and value multiset while
    removing the original temporal order.

    \item \textbf{Flat} replaces the observed values in each series by
    their mean, approximately preserving the observed level while removing
    within-series variation.

    \item \textbf{Noise} replaces the observed values with independent
    Gaussian draws parameterized by the series' observed mean and standard
    deviation.

    \item \textbf{Swap} replaces the numerical series with a compatible
    series from another sample. 

    \item \textbf{Answer-only} removes the numerical series while retaining
    the question and all other non-series information.

    \item \textbf{Zero} replaces the numerical values by zero while
    preserving the series length and input structure.
\end{itemize}

For sample \(i\), we use the notation $z_i=(x_i,q_i,y_i)$,
where \(x_i\) denotes the numerical series, \(q_i\) contains the
question and other non-series information, and \(y_i\) is the
source-defined Original target. Let \(T_c\) denote the numerical
transformation for condition \(c\), with \(T_0(x_i)=x_i\) for
original, and let \(R_s\) be the input renderer for system \(s\).
The prediction under condition \(c\) is
\[
\hat{y}_{si}^{(c)}
=
f_s\!\left(
R_s\!\left(T_c(x_i),q_i\right)
\right).
\tag{1}
\]

Note that \(\hat{y}_{si}^{(c)}\) is evaluated against the Original target \(y_i\). 
Appendix~\ref{app:operators} specifies the transformation details, masking rules, degenerate cases, numerical rendering, and Swap donor selection procedure.



\subsection{Metrics}

For forecasting, we report mean absolute scaled error (MASE)
\citep{hyndman2006accuracy}. 
Lower MASE indicates better forecasting performance.

For anomaly detection, we report binary accuracy over the normal and anomalous labels. Pattern recognition, causal diagnosis, and decision intervention are evaluated by accuracy. Cross-series comparison uses relation accuracy, which counts a prediction as correct when its canonical relation matches the target relation.

\section{Experiments}
\label{sec:results}

\subsection{Does specialization improve native task performance? (RQ1)}
\label{sec:native-results}

We consider three systems with released evaluation data: \tomni{}, \chatts{}, and \tovl{}. We first reproduce their reported evaluations before comparing each specialized system with its matched backbone. The reproduced results are close to the reported values on most native metrics. \chatts{} differs by at most 4.0\% across its six reported metrics, while six of the eight \tomni{} metrics differ by at most 3.3\%. The larger reproduction discrepancies occur in forecasting, particularly for \tomni{} on the out-of-distribution (OOD) split. Full reproduction results are reported in Appendix~\ref{app:native-compare}.

\captionsetup[subfigure]{
    font=footnotesize,
    justification=centering,
    singlelinecheck=true,
    skip=2pt
}

\begin{figure*}[t]
\centering
\begin{subfigure}[t]{0.325\textwidth}
    \centering
    \includegraphics[width=1.05\linewidth]
        {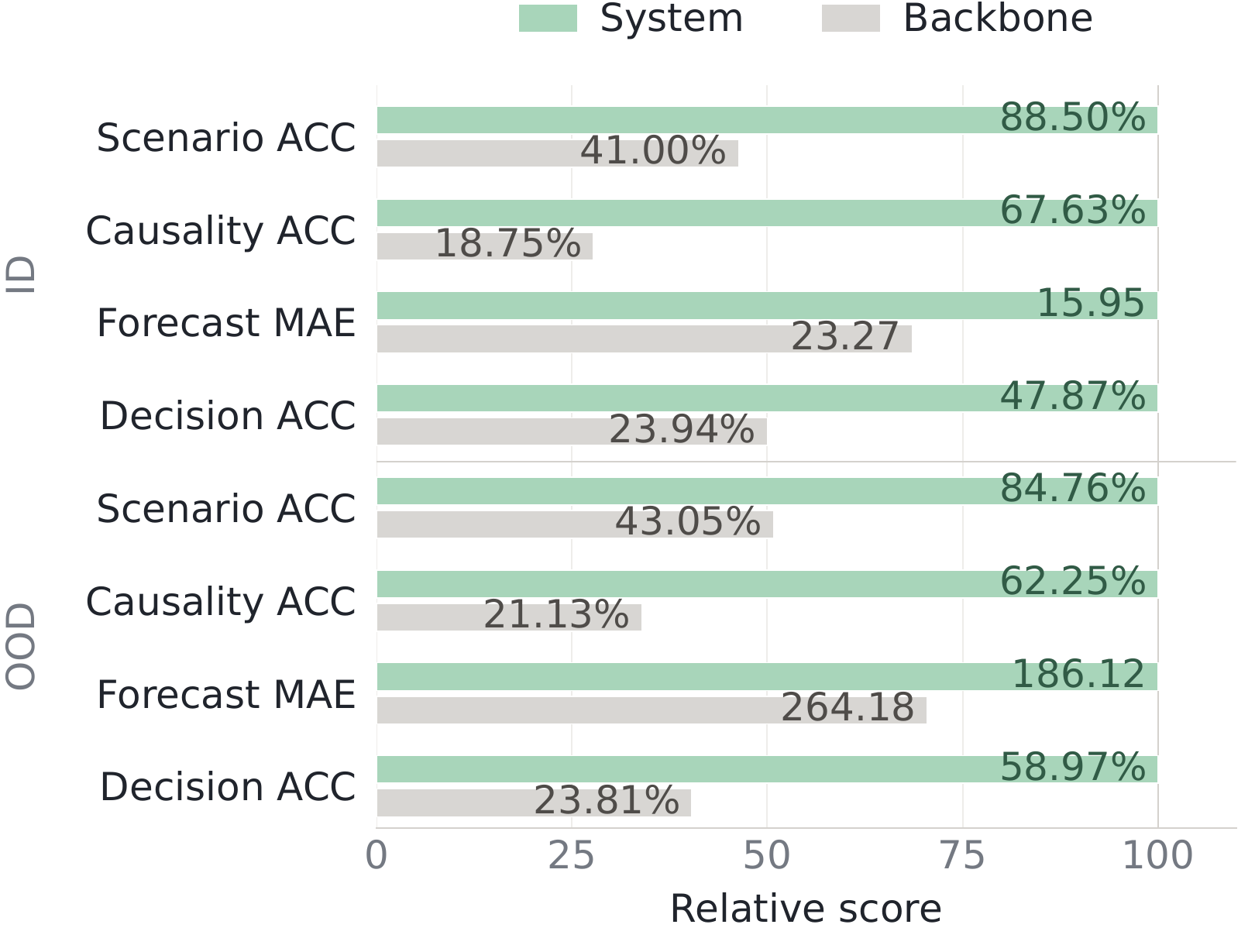}
    \caption{TimeOmni-1 vs.\ Qwen2.5-7B}
    \label{fig:backbone-timeomni1}
\end{subfigure}
\hfill
\begin{subfigure}[t]{0.325\textwidth}
    \centering
    \includegraphics[width=1.05\linewidth]
        {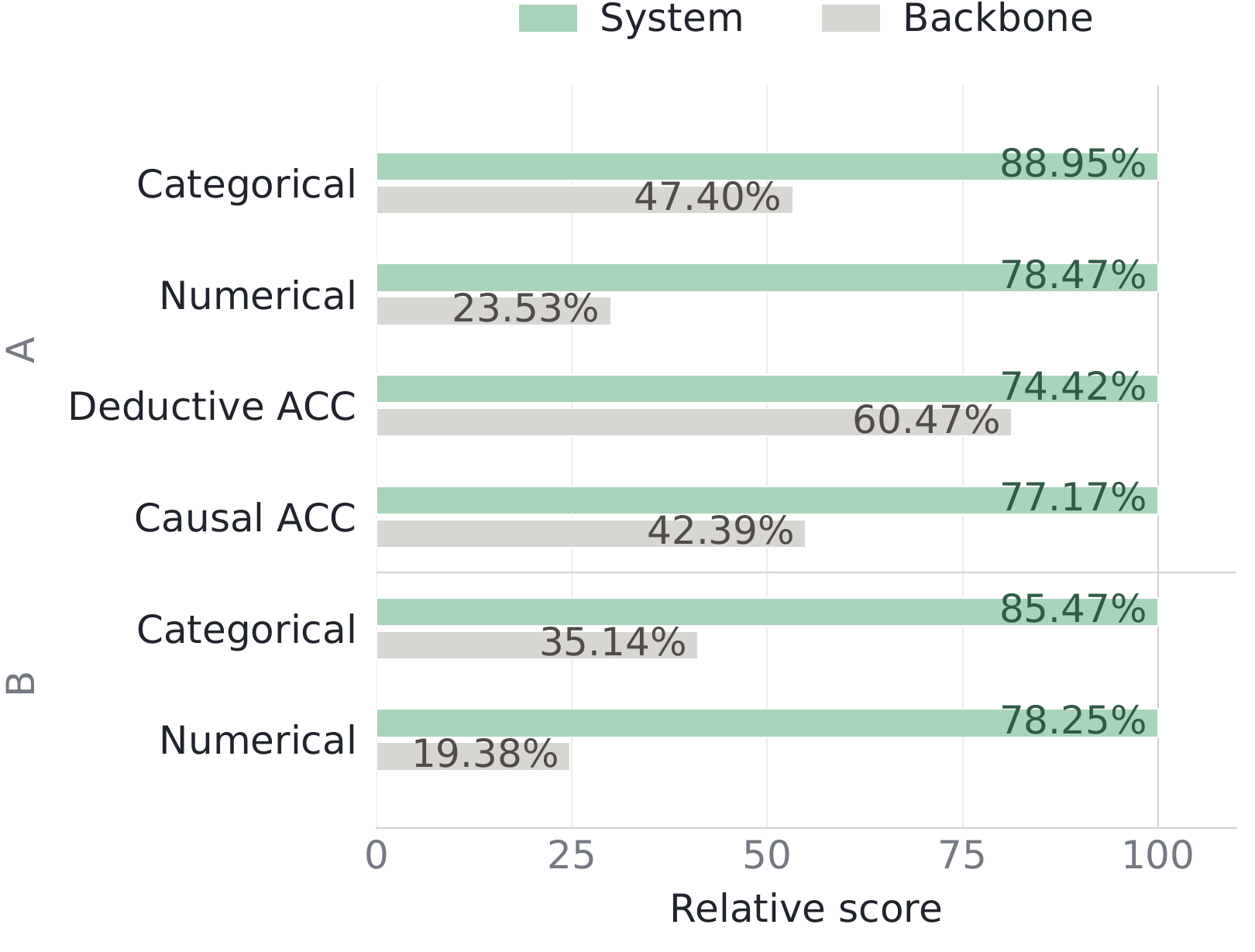}
    \caption{ChatTS vs.\ Qwen2.5-14B}
    \label{fig:backbone-chatts}
\end{subfigure}
\begin{subfigure}[t]{0.325\textwidth}
    \centering
    \includegraphics[width=1.05\linewidth]
        {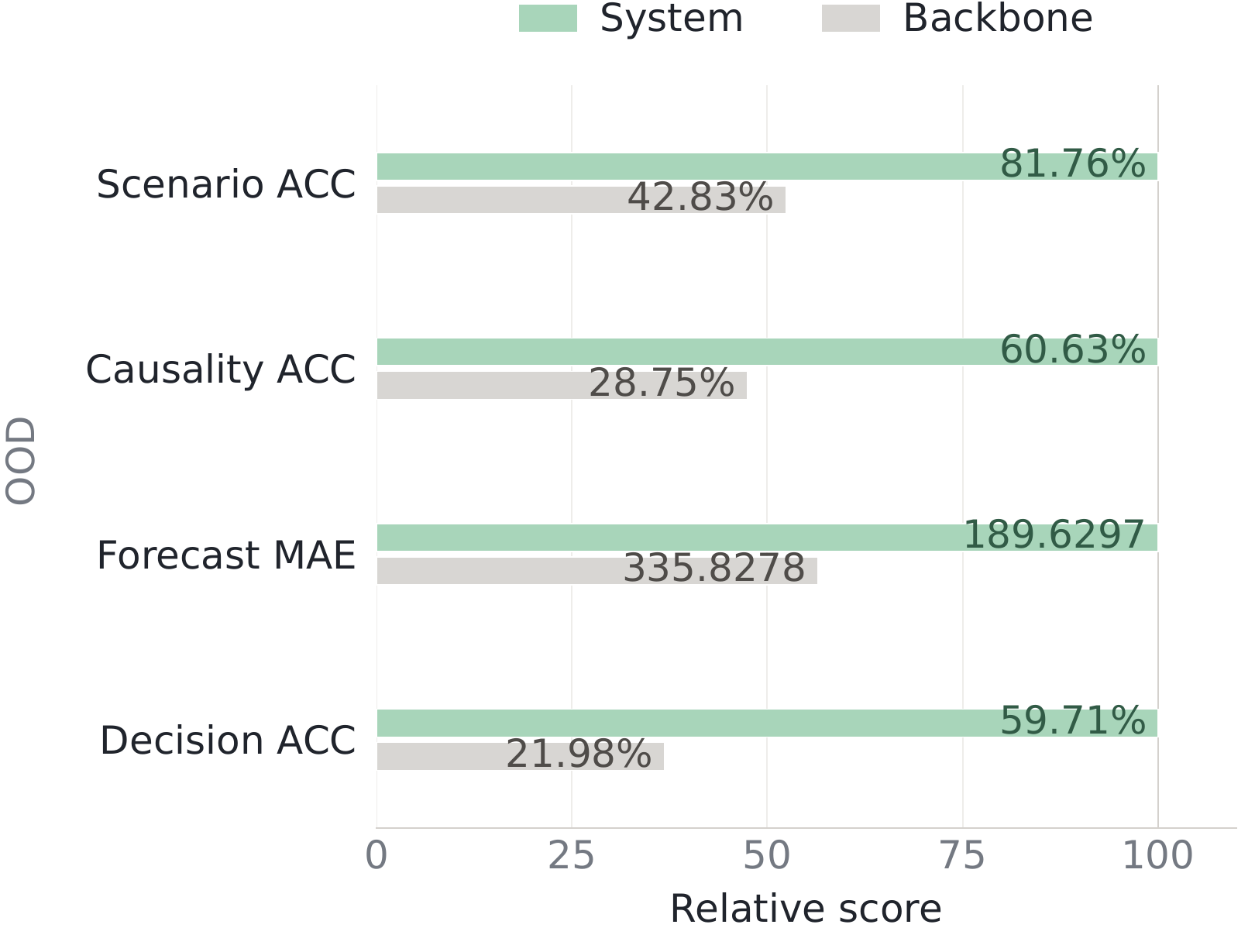}
    \caption{TimeOmni-VL vs.\ BAGEL-7B}
    \label{fig:backbone-timeomnivl}
\end{subfigure}
\caption{Systems and matched backbones on native evaluations. Bars are normalized separately within each task so that the
better result equals 100; bar heights are intended only for within-task comparison. Labels report the original metric values.}
\label{fig:system-backbone-comparison}
\end{figure*}

The specialization gains are consistent across the available native comparisons. As shown in Figure~\ref{fig:system-backbone-comparison}, \tomni{} outperforms Qwen2.5-7B~\citep{qwen2025qwen25technicalreport} across all of its reported ID and OOD tasks, including both accuracy-based tasks and forecasting. \chatts{} likewise improves over Qwen2.5-14B~\citep{qwen2025qwen25technicalreport} on all six reported metrics, and \tovl{} improves over BAGEL-7B~\citep{deng2025bagel} on all four OOD metrics.
These results show that specialization consistently improves native task performance for the systems that can be compared with their matched backbones. 
However, the comparison does not identify how the numerical series contributes to those predictions. A specialized system may achieve higher native performance without every prediction depending on the sample-specific temporal evidence in the same way.

\subsection{How Do Systems Respond to Changes in Numerical Evidence? (RQ2)}
\label{sec:rq2}

\begin{table*}[t]
\centering
\scriptsize
\caption{Scores of four time-series QA systems under original and six interventions. All conditions are scored against the original ground-truth answer. Bold indicates the best score. ``--'' indicates that the system produced no scorable forecasting outputs.}
\label{tab:main-results}
\setlength{\tabcolsep}{3.8pt}
\renewcommand{\arraystretch}{1.10}
\resizebox{\textwidth}{!}{%
\begin{tabular}{lllccccccc}
\toprule
Task & Metric & Model
& Original & Shuffle & Flat & Noise & Swap & Answer-only & Zero \\
\midrule

\multirow{4}{*}{Forecasting}
& \multirow{4}{*}{MASE $\downarrow$}
& TimeOmni-1
& 1.739 & 1.929 & \textbf{1.321} & 2.060
& 26.848 & 17.418 & 7.235 \\

& & ChatTS
& -- & -- & -- & --
& -- & -- & -- \\

& & Time-MQA
& 2.258 & 2.287 & \textbf{2.106} & 2.260
& 23.380 & 38.241 & 12.706 \\

& & TimeOmni-VL
& 2.650 & 3.001 & \textbf{2.604} & 3.291
& 36.109 & 38.546 & 13.221 \\

\midrule

\multirow{4}{*}{Anomaly detection}
& \multirow{4}{*}{Accuracy (\%) $\uparrow$}
& TimeOmni-1
& \textbf{60.70} & 55.44 & 45.61 & 46.32
& 45.96 & 45.61 & 45.61 \\

& & ChatTS
& 62.46 & \textbf{68.88} & 45.61 & 58.60
& 54.39 & 54.39 & 45.61 \\

& & Time-MQA
& 64.91 & \textbf{65.46} & 45.61 & 56.14
& 52.28 & 45.61 & 45.61 \\

& & TimeOmni-VL
& \textbf{70.88} & 56.49 & 46.32 & 46.67
& 47.02 & 45.61 & 45.61 \\

\midrule

\multirow{4}{*}{Pattern recognition}
& \multirow{4}{*}{Accuracy (\%) $\uparrow$}
& TimeOmni-1
& \textbf{60.51} & 47.67 & 26.00 & 44.14
& 52.01 & 52.81 & 25.52 \\

& & ChatTS
& \textbf{59.91} & 48.15 & 23.76 & 45.75
& 59.07 & 59.23 & 23.76 \\

& & Time-MQA
& \textbf{55.06} & 51.85 & 33.07 & 50.08
& 52.65 & 43.98 & 33.23 \\

& & TimeOmni-VL
& 58.35 & 50.40 & 18.62 & 51.20
& 56.02 & \textbf{58.91} & 19.42 \\

\midrule

\multirow{4}{*}{Cross-series comparison}
& \multirow{4}{*}{Accuracy (\%) $\uparrow$}
& TimeOmni-1
& 58.79 & 57.06 & 35.78 & \textbf{59.41}
& 55.51 & 50.87 & 38.49 \\

& & ChatTS
& \textbf{54.93} & 49.32 & 31.91 & 48.74
& 46.86 & 46.23 & 26.31 \\

& & Time-MQA
& 46.81 & 44.49 & 29.01 & \textbf{47.64}
& 40.23 & 34.82 & 25.53 \\

& & TimeOmni-VL
& 53.00 & \textbf{53.31} & 26.89 & 44.68
& 35.20 & 35.59 & 19.73 \\

\midrule

\multirow{4}{*}{Causal diagnosis}
& \multirow{4}{*}{Accuracy (\%) $\uparrow$}
& TimeOmni-1
& 33.67 & 31.90 & 30.89 & \textbf{33.92}
& 25.32 & 27.85 & 21.27 \\

& & ChatTS
& 28.35 & 28.86 & 27.09 & 29.11
& 30.63 & \textbf{31.39} & 28.86 \\

& & Time-MQA
& 27.19 & 25.57 & 24.56 & \textbf{27.34}
& 23.29 & 19.24 & 16.96 \\

& & TimeOmni-VL
& \textbf{33.42} & 29.87 & 25.82 & 30.13
& 26.58 & 29.62 & 19.75 \\

\midrule

\multirow{4}{*}{Decision intervention}
& \multirow{4}{*}{Accuracy (\%) $\uparrow$}
& TimeOmni-1
& \textbf{43.43} & 23.43 & 26.29 & 21.14
& 34.29 & 21.14 & 29.16 \\

& & ChatTS
& 26.14 & 21.71 & 28.57 & 24.57
& 24.57 & 26.29 & \textbf{29.71} \\

& & Time-MQA
& 26.17 & 24.00 & \textbf{27.43} & 20.57
& 19.43 & 22.29 & 20.57 \\

& & TimeOmni-VL
& \textbf{49.71} & 23.43 & 30.29 & 30.86
& 42.86 & 27.43 & 26.29 \\

\bottomrule
\end{tabular}
}
\end{table*}

Table~\ref{tab:main-results} shows that \Original{} is not uniformly the best-performing condition. \Flat{} achieves the lowest forecasting MASE, whereas the effects on discrete tasks vary across systems and interventions. To interpret these differences, we examine whether similar scores hide different predictions, whether removing or simplifying the series induces a recognizable fallback, and whether the correct answer itself changes with the numerical evidence.

\paragraph{Similar accuracy can hide different predictions.}
To analyze sample-level sensitivity, we compare the semantic answers produced under Original and Answer-only. Correct-to-wrong (C$\rightarrow$W) and wrong-to-correct (W$\rightarrow$C) record changes in correctness against the original ground-truth answer. Changed records any difference between the two semantic answers, including a change from one incorrect answer to another.
Table~\ref{tab:rq2-ao} shows that \chatts{} and \tovl{} retain similar pattern-recognition accuracy after the series is removed, despite changing many individual predictions. Their C$\rightarrow$W and W$\rightarrow$C transitions largely offset each other on the jointly valid samples.
\begin{wraptable}{r}{0.44\linewidth}
\centering
\scriptsize
\caption{Pattern-recognition transitions from Original to Answer-only.}
\label{tab:rq2-ao}
\setlength{\tabcolsep}{4.0pt}
\renewcommand{\arraystretch}{1.00}
\begin{tabular}{lccc}
\toprule
System
& C$\rightarrow$W
& W$\rightarrow$C
& Changed \\
& (\%)
& (\%)
& (\%)\\
\midrule
TimeOmni-1  
& 24.11 & 16.34 & 48.22  \\
ChatTS      
& 14.13 & 12.84 & 30.50  \\
Time-MQA    
& 17.62 & 11.88 & 34.65  \\
TimeOmni-VL 
& 19.69 & 20.38 & 47.95  \\
\bottomrule
\end{tabular}
\end{wraptable}
Thus, similar aggregate scores can conceal changes in which samples are answered correctly. The predictions are sensitive to the numerical input, but this sensitivity alone does not establish that the input is used appropriately. 
Appendix~\ref{app:rq2-transitions} reports the other discrete tasks.

\paragraph{Answer-only induces single-class fallback in anomaly detection.}
We inspect the predicted label distribution to determine whether the
remaining Answer-only accuracy reflects discrimination between samples.
All four systems predict a single class: \tomni{}, \timemqa{}, and
\tovl{} label every sample as \emph{normal}, whereas \chatts{} labels every sample as \emph{anomalous}. A constant prediction is correct whenever a sample belongs to the selected class, so its accuracy depends only on that class's frequency. The residual Answer-only score reflects a fixed-class fallback rather than retained discrimination between normal and anomalous samples.

\begin{figure*}[t]
\centering
\begin{minipage}[t]{0.47\textwidth}
    \centering
    \includegraphics[width=0.7\linewidth]
    {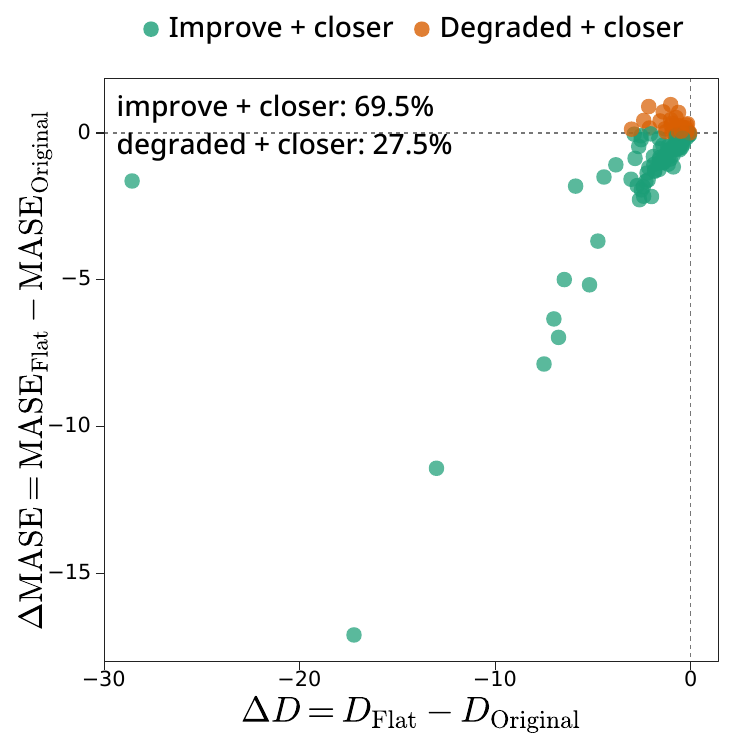}

    {\small (a) \tomni{} original and flat forecasting results.}
\end{minipage}
\hfill
\begin{minipage}[t]{0.47\textwidth}
    \centering
    \includegraphics[width=0.7\linewidth]
    {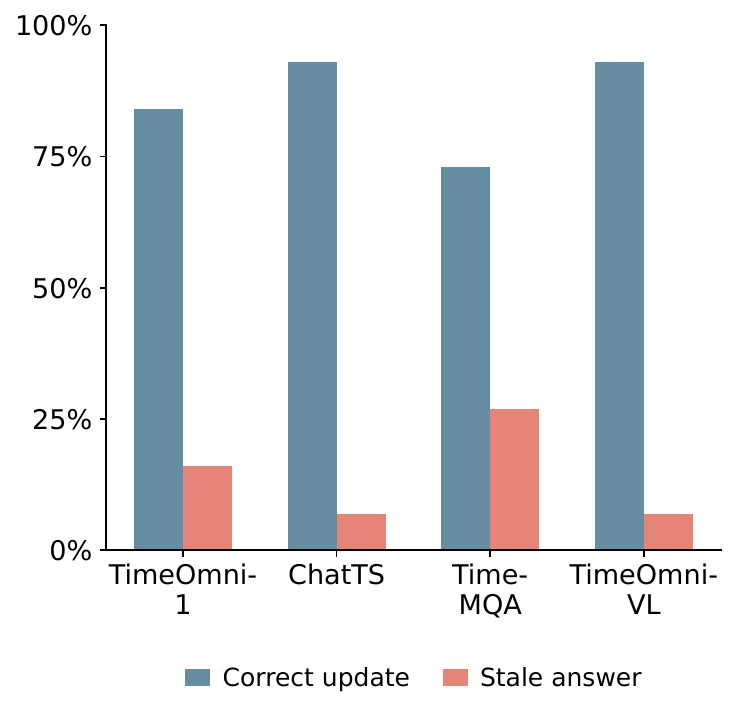}

    {\small (b) Correct updates vs. stale answers.}
\end{minipage}

\caption{Paired diagnostics of intervention responses.
(\textbf{a}) Each point is a forecasting sample scorable under both Original and Flat. The horizontal axis is
$\Delta D=D^{\mathrm{Flat}}-D^{\mathrm{Original}}$, the change in scaled distance to the fixed Original-history-mean forecast; the vertical axis is
$\Delta\mathrm{MASE}
=\mathrm{MASE}^{\mathrm{Flat}}-\mathrm{MASE}^{\mathrm{Original}}$.
Improve + closer denotes $\Delta D<0$ and $\Delta\mathrm{MASE}<0$;
Degraded + closer denotes $\Delta D<0$ and $\Delta\mathrm{MASE}>0$.
(\textbf{b}) On cross-series sample--intervention pairs whose correct answer changes, the bars show correct updates and stale answers. Correct updates match the recomputed target, and stale answers match the original ground-truth answer.}
\label{fig:rq2-diagnostics}
\end{figure*}

\paragraph{Flat shifts forecasts toward a history-mean reference.}
To understand the forecasting behavior associated with Flat, we compare Original and Flat on samples that are scorable under both conditions.
We use a constant forecast equal to the mean of the Original history as a fixed reference. Let $D$ denote the mean absolute distance from a model forecast to this reference, divided by the same Original-history scale used for MASE. Unlike MASE, which measures error against the future observations, $D$ measures similarity to the reference forecast.
Figure~\ref{fig:rq2-diagnostics}(a) plots the changes in these two quantities: $\Delta D<0$ indicates movement toward the reference, and $\Delta\mathrm{MASE}<0$ indicates improved forecasting.

Flat moves nearly all paired forecasts toward the history-mean reference. For TimeOmni-1 and Time-MQA, this movement is more often accompanied by lower error; for TimeOmni-VL, it is more often accompanied by higher error. The common response is therefore a shift toward a simple level forecast, not a uniform improvement in prediction. This distinction helps explain why removing temporal variation can improve some forecasts without demonstrating more effective modeling of the original temporal structure. Full paired results and an alternative last-value reference are provided in Appendix~\ref{app:rq2-flat}.

\paragraph{Errors against the original answer can include appropriate updates.}
An intervention can change the answer implied by the input. For instance, flattening two series makes both ranges zero, so a range-comparison question may no longer have its original answer. We therefore recompute answers for the subset of cross-series questions whose targets can be determined unambiguously from the transformed series.
For sample--intervention pairs whose correct answer changes, a \emph{correct update} matches the recomputed answer, whereas a \emph{stale answer} matches the original ground-truth answer.
These labels refer to the two targets, not to whether the system repeats its own Original prediction. Figure~\ref{fig:rq2-diagnostics}(b) compares their relative shares within these two categories; other errors are reported separately in Appendix~\ref{app:rq2-targetaware}.

Correct updates are more frequent than stale answers for every system, and scoring against the recomputed answers increases accuracy on the eligible subset. The distinction is particularly relevant to Flat and Zero, which change the target much more often than Shuffle in this subset. Some errors against the original answer therefore reflect appropriate responses to the transformed evidence rather than a loss of task ability. However, stale answers and other errors remain, showing that appropriate updating is not consistent across all samples and
interventions.

\subsection{Where Does the Evidence-to-Answer Reasoning Path Fail? (RQ3)}
\label{sec:rq3}

In this section, we evaluate the visible rationales of TimeOmni-1 and TimeOmni-VL under Original using GPT-5.6 Sol\footnote{Accessed via the OpenAI API between July 19 and September 24, 2026.}~\citep{openai2026gpt56} as the LLM judge, without providing the ground-truth answer. We first examine individual numerical claims and then distinguish factual grounding, inference validity, and conclusion consistency at the output level. Appendix~\ref{app:judge} provides the factual-grounding prompt, along with the protocol and results of an independent human validation of the LLM-based adjudication.

\begin{figure*}[t]
\centering
\captionsetup[subfigure]{font=scriptsize,skip=0pt}
\begin{subfigure}[t]{0.47\textwidth}
    \centering
    \includegraphics[width=0.8\linewidth]{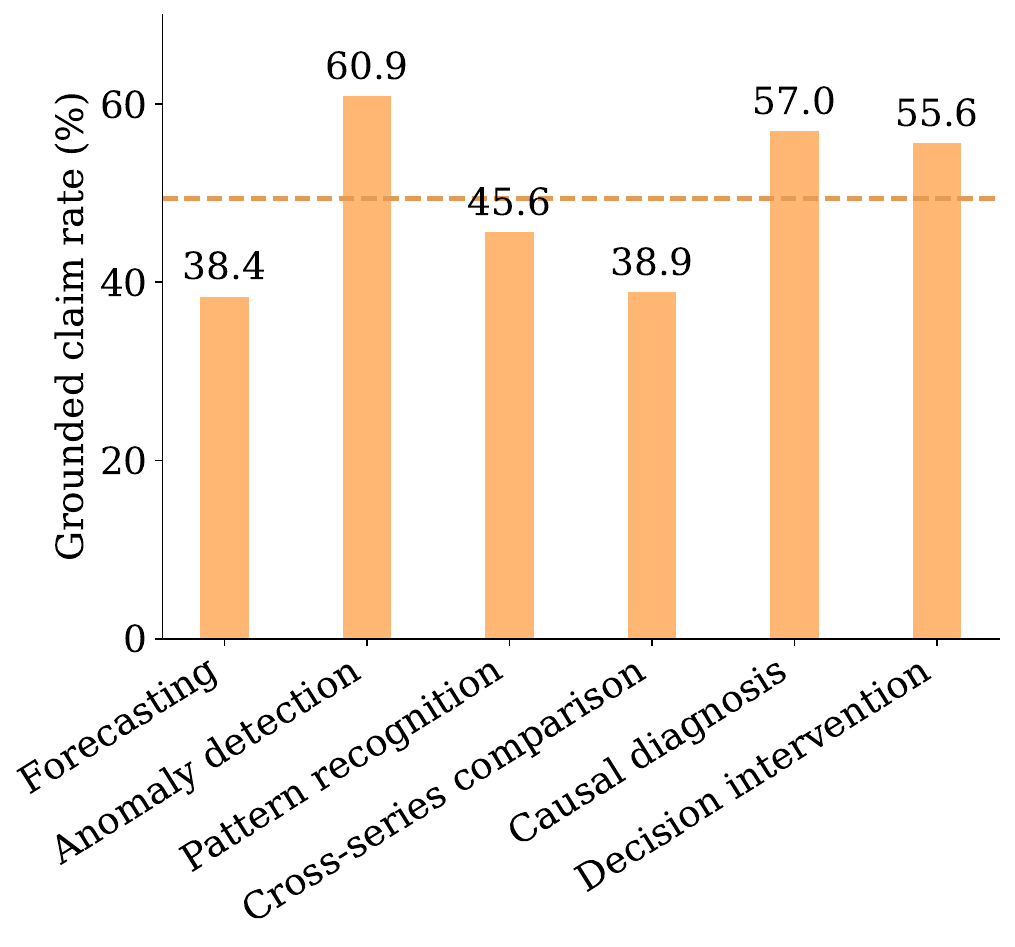}
    \caption{\tomni{}}
    \label{fig:timeomni-grounding-task}
\end{subfigure}
\hfill
\begin{subfigure}[t]{0.47\textwidth}
    \centering
    \includegraphics[width=0.8\linewidth]{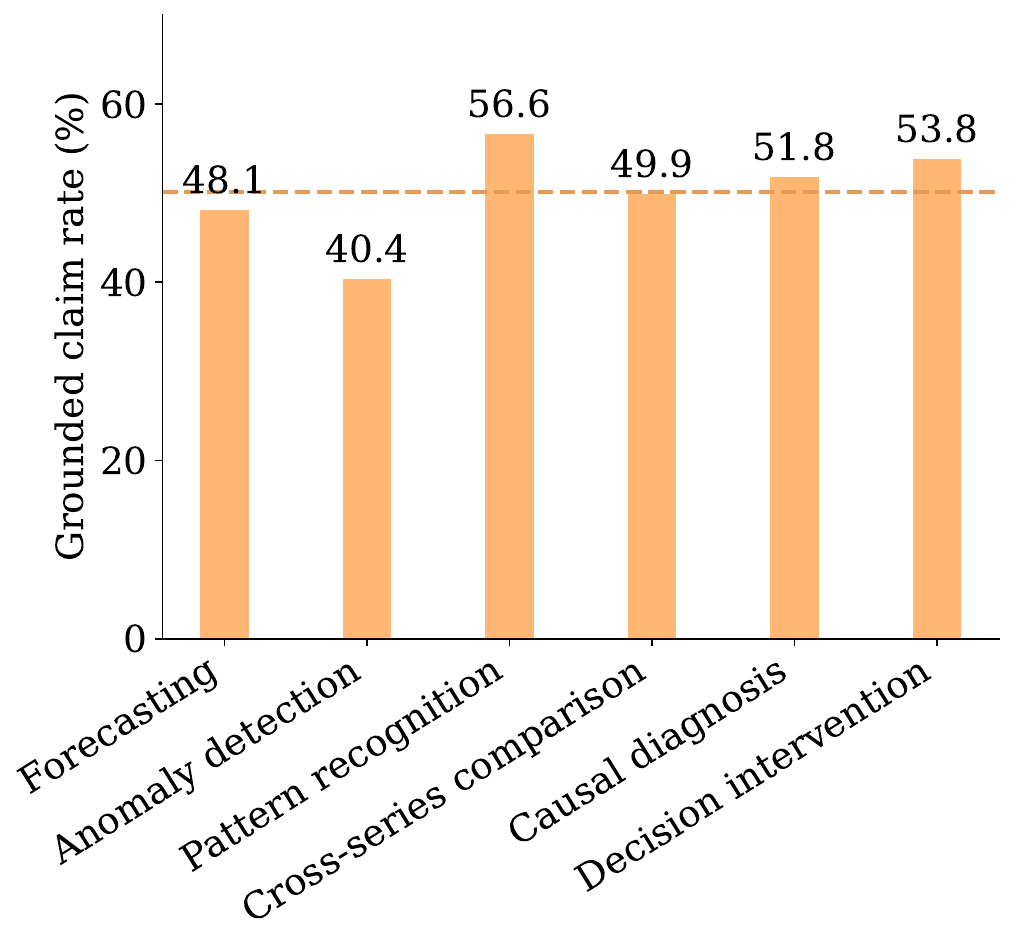}
    \caption{\tovl{}}
    \label{fig:timeomnivl-grounding-task}
\end{subfigure}
\caption{Grounded claim rate across six tasks under \Original{}. Dashed lines indicate the overall rates, 49.4\% for \tomni{}, and 50.1\% for \tovl{}.}
\label{fig:rationale-grounding}
\end{figure*}

\paragraph{Numerical claims are only partially grounded in the input.}
We extract atomic claims about the observed time series and check them against the supplied numerical evidence. Grounded claims are judged supported by that evidence; the remaining labels distinguish contradicted statements from statements that cannot be verified.
Figure~\ref{fig:rationale-grounding} reports the grounded claim rates across tasks. Only about half of the extracted claims are supported for either system, but similar overall rates conceal different task-level weaknesses.
TimeOmni-1 has lower grounding rates in forecasting and cross-series comparison than in anomaly detection, whereas TimeOmni-VL has its lowest rate in anomaly detection. Thus, neither system provides uniformly reliable numerical descriptions across tasks. 


\paragraph{Conclusion consistency does not guarantee grounded or valid reasoning.}
We evaluate the rationale beyond its individual claims. Factual grounding checks whether the stated premises are supported by the numerical input under the strict output-level criterion. Inference validity asks whether all assessable key reasoning steps are valid given the stated premises, even when those premises are factually incorrect. Conclusion consistency checks whether the rationale's conclusion agrees with the system's final answer, not whether that answer is correct.

Table~\ref{tab:reasoning-audit} shows low pass rates under the strict grounding criterion, and fewer than half of the outputs pass inference validity for either system. Conclusion consistency is substantially higher, especially for TimeOmni-VL. These measures capture different requirements: supported individual statements do not establish a grounded rationale, and agreement with the final answer does not establish that the preceding inference is valid.

Case~1 in Appendix~\ref{app:cases} illustrates the distinction. The rationale correctly reports amplitudes of $14$ and $7.29$, but concludes that the latter is larger; its final answer follows that mistaken comparison. The relevant numerical values are correct and the conclusion agrees with the answer, but the inference is invalid.
The quantitative results and cases show why checking only the final conclusion overlooks failures in describing the evidence and
reasoning from it.

\begin{table}[t]
\centering
\small
\caption{Output-level rationale audit under original setting. Values are pass rates (\%) for the reported factual-grounding, inference-validity, and conclusion-consistency criteria. Conclusion consistency concerns agreement with the system's own final answer.}
\label{tab:reasoning-audit}
\setlength{\tabcolsep}{5pt}
\renewcommand{\arraystretch}{1.05}
\begin{tabular}{lccc}
\toprule
 & Factual grounding & Inference validity &
Conclusion consistency  \\
\midrule
\tomni{} & 2.1 & 27.3 & 64.0 \\
\tovl{}  & 11.2 & 34.3 & 92.2 \\
\bottomrule
\end{tabular}
\end{table}

\section{Discussion}
\label{sec:discussion}

We relate the observed behavior to each system's public training design and available checkpoints. The analysis identifies limitations that remain after specialization, rather than attributing errors to a
particular training choice. Appendix~\ref{app:training-diagnostics}
documents the public evidence and release constraints; a training signal that is not documented is not assumed to be absent.

\paragraph{TimeOmni-1: task-level optimization does not ensure grounded intermediate reasoning.}
TimeOmni-1 combines human-guided chain-of-thought (CoT) supervision with reinforcement learning on forecasting, causal, and decision tasks. Its native performance gains therefore coexist with the grounding and inference failures identified in RQ3. This contrast shows that task and CoT supervision can improve final-answer performance without ensuring that the numerical statements and subsequent inferences in a rationale are reliable. The public objectives emphasize task-level outcomes, leaving intermediate numerical verification as a direction for further study.

\paragraph{ChatTS: broad task coverage does not ensure reliable evidence-responsive answers.}
Our review of the ChatTS corpus identifies multi-series,
cross-series, causal, and rationale-like supervision (Appendix~\ref{app:chatts-corpus}). The observed errors therefore occur within task families represented in the public training data, rather than only on wholly unrepresented types of questions. RQ2 also shows both appropriate updates and incorrect responses after the evidence changes.
These observations motivate examining how training connects numerical evidence to the required answer, rather than treating broad task coverage as sufficient evidence of reliable numerical reasoning.

\paragraph{Time-MQA: adaptation changes output validity and correctness unevenly.}
\begin{wrapfigure}{r}{0.48\linewidth}
\centering\includegraphics[width=1.0\linewidth]{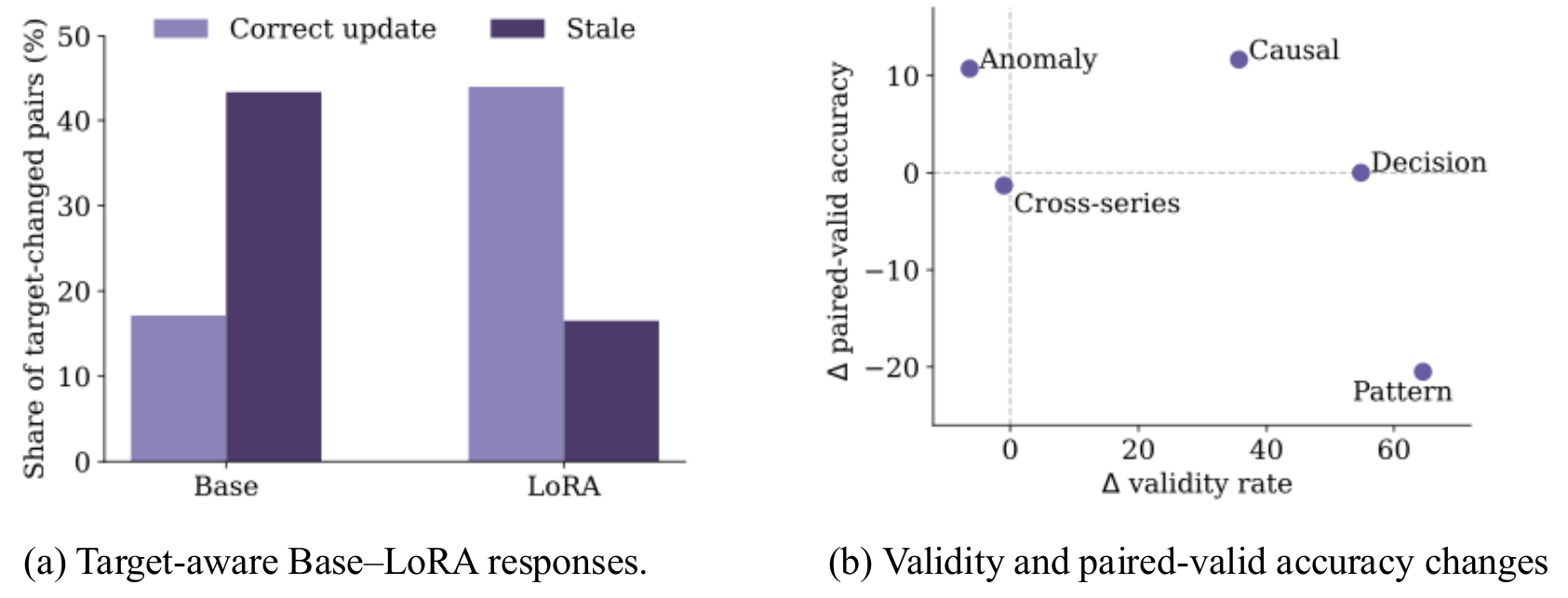}
\caption{\textbf{(a)} Correct updates and stale answers on 327 cross-series pairs whose targets change after intervention; other errors are reported in Appendix~\ref{app:timemqa-contrast}.
\textbf{(b)} Base$\rightarrow$LoRA changes in output validity and paired-valid accuracy across discrete tasks.}
\label{fig:timemqa-training-diagnostic}
\end{wrapfigure}
The released LoRA adapter allows a comparison between a fixed public base and the same base with specialization applied.
Figure~\ref{fig:timemqa-training-diagnostic}(a) shows more correct updates and fewer stale answers on target-changed cross-series pairs after adaptation. The gains are not uniform across interventions: the largest increases in correct updates occur under Flat and Zero.
Figure~\ref{fig:timemqa-training-diagnostic}(b) further separates output validity, the fraction of outputs with recoverable semantic answers, from paired-valid accuracy, which is measured on samples with valid answers from both checkpoints. In pattern recognition, adaptation substantially increases validity but lowers paired-valid accuracy; in causal diagnosis, both improve. Thus, an aggregate accuracy gain can reflect different changes in output validity and correctness across tasks. The comparison uses matched inputs, decoding, and scoring, but is a checkpoint contrast rather than an exact training-stage ablation. Details are provided in Appendix~\ref{app:timemqa-contrast}.


\paragraph{TimeOmni-VL: answer consistency is narrower than reasoning reliability.}
Despite its CoT-based training, TimeOmni-VL exhibits a substantial gap between conclusion consistency and the grounding and inference criteria in RQ3. Therefore, aligning the rationale's conclusion with the final answer is a narrower success than providing a numerically supported and valid derivation. 


These findings distinguish three requirements that specialization need not improve together: producing valid answers, responding appropriately to changed numerical evidence, and providing grounded, logically valid rationales.

\section{Conclusion}
\label{sec:conclusion}

We evaluate four time-series QA systems through native reproduction, numerical-evidence interventions, and visible-rationale audits. Specialization improves native task performance, but this does not establish reliable use of numerical evidence or reliable visible reasoning. Similar task-level scores can conceal substantial prediction turnover, while intervention gains can accompany shifts toward simple forecasting references. When the numerical input changes the correct answer, scoring against the original target can also penalize appropriate updates. The rationale audits further distinguish supported numerical claims, valid inference, and consistency with the final answer.
These findings show that predictive performance, evidence-responsive behavior, and rationale reliability should be evaluated separately.

\bibliographystyle{iclr2027_conference}
\bibliography{references}

\appendix
\section{Native Evaluation and Reproduction}
\label{app:native}

\subsection{Systems}

We evaluate four time-series language models. Table~\ref{tab:systems_all} summarizes their backbones, time-series inputs, training procedures, and native evaluation sets.

\paragraph{TimeOmni-1~\citep{guan2026timeomni1}.}
It is built on Qwen2.5-7B-Instruct~\citep{qwen2025qwen25technicalreport}. It is first supervised fine-tuned on chain-of-thought samples from TSR-Suite and then optimized with Group Relative Policy Optimization (GRPO) using task-specific rewards.

\paragraph{ChatTS~\citep{xie2025chatts}.}
ChatTS uses Qwen2.5-14B-Instruct~\citep{qwen2025qwen25technicalreport} as its language backbone. Training uses synthetic time-series--text data: alignment pre-training first connects the time-series encoder to the language model, followed by supervised fine-tuning on time-series understanding and reasoning tasks.

\paragraph{TimeOmni-VL~\citep{guan2026timeomnivl}.}
TimeOmni-VL is built on BAGEL-7B-MoT~\citep{deng2025bagel}. The pretrained BAGEL checkpoint is post-trained on TSUMM-Suite with chain-of-thought supervision for time-series understanding and generation.

\paragraph{Time-MQA~\citep{kong2025timemqa}.}
Time-MQA uses Qwen2.5-7B~\citep{qwen2025qwen25technicalreport} as its backbone. The released Qwen variant is trained on the TSQA corpus with LoRA, covering forecasting, imputation, anomaly detection, classification, and reasoning tasks.


\begin{table*}[h]
\centering
\small
\caption{Systems and original evaluation settings.}
\label{tab:systems_all}
\setlength{\tabcolsep}{7pt}
\begin{tabular}{lllll}
\toprule
System & Backbone & TS input & Training & Evaluation data \\
\midrule

\tomni{}
& Qwen2.5-7B
& Text
& CoT-SFT $\rightarrow$ GRPO
& \tsr{} \\

\chatts{}
& Qwen2.5-14B
& TS encoder
& Alignment $\rightarrow$ SFT
& Eval A/B \\

\tovl{}
& BAGEL-7B
& TS image
& CoT post-training
& \tsr{} (OOD) \\

\timemqa{}
& Qwen2.5-7B
& Text + context
& Continual PT (LoRA)
& TSQA \\


\bottomrule
\end{tabular}
\end{table*}

\subsection{Native comparison}
\label{app:native-compare}


\paragraph{Reproduction.}
In Table~\ref{tab:reproduction-a}, we compare our results with the reported results. We evaluate \tomni{} on the ID and OOD splits of TSR-Suite. We report the objective metrics from Evaluation Sets A/B for \chatts. For \tovl{}, we follow the reasoning route on TSR-Suite-OOD.


\begin{table*}[t]
\centering
\footnotesize
\setlength{\tabcolsep}{4pt}
\caption{Reproduction results. For MAE, lower is better; all other metrics are higher-is-better.}
\label{tab:reproduction-a}
\begin{tabularx}{\textwidth}{@{}llYccc@{}}
\toprule
System & Split & Task/metric & Reproduced & Paper & Difference \\
\midrule
\multirow{8}{*}{\tomni{}}
& ID & Scenario ACC & 88.50 & 90.70 & $-2.20$ \\
& ID & Causality ACC & 67.60 & 69.30 & $-1.70$ \\
& ID & Forecasting MAE & 15.95 & 14.30 & $+1.65$ \\
& ID & Decision ACC & 47.90 & 47.90 & $0.00$ \\
& OOD & Scenario ACC & 84.80 & 87.70 & $-2.90$ \\
& OOD & Causality ACC & 62.30 & 64.00 & $-1.70$ \\
& OOD & Forecasting MAE & 186.12 & 145.53 & $+40.59$ \\
& OOD & Decision ACC & 59.00 & 58.90 & $+0.10$ \\
\midrule
\multirow{6}{*}{\chatts{}}
& A & Overall categorical & 0.890 & 0.889 & $+0.001$ \\
& A & Overall numerical & 0.785 & 0.788 & $-0.003$ \\
& A & Deductive reasoning & 0.744 & 0.744 & $0.000$ \\
& A & Causal reasoning & 0.772 & 0.804 & $-0.032$ \\
& B & Overall categorical & 0.855 & 0.862 & $-0.007$ \\
& B & Overall numerical & 0.783 & 0.787 & $-0.004$ \\
\midrule
\multirow{4}{*}{\tovl{}}
& OOD & Scenario ACC & 81.80 & 84.00 & $-2.20$ \\
& OOD & Causality ACC & 60.60 & 61.30 & $-0.70$ \\
& OOD & Forecasting MAE & 189.63 & 163.79 & $+25.84$ \\
& OOD & Decision ACC & 59.70 & 61.40 & $-1.70$ \\
\bottomrule
\end{tabularx}
\end{table*}

\paragraph{Comparison with matched backbone.}
We evaluate each matched backbone on the same tasks in Table~\ref{tab:native_backbone_comparison}. 
These comparisons establish the native specialization gains but do not determine how predictions depend on the numerical series of individual samples.

\begin{table*}[t]
\centering
\small
\caption{
Performance of time-series QA systems and their matched
backbones. Accuracy and score metrics are reported as percentages, forecasting MAE retains its original scale. 
}
\label{tab:native_backbone_comparison}
\setlength{\tabcolsep}{6.5pt}
\renewcommand{\arraystretch}{1.06}

\begin{tabular}{@{}lllc cc@{}}
\toprule
System / backbone & Split & Task / metric
& System & Backbone & Improvement \\
\midrule

\multirow{8}{*}{\tomni{} / Qwen2.5-7B}
& \multirow{4}{*}{ID}
& Scenario ACC     & 88.50  & 41.00  & +47.50 \\
&& Causality ACC   & 67.63  & 18.75  & +48.88 \\
&& Forecasting MAE & 15.95  & 23.27  & +7.32  \\
&& Decision ACC    & 47.87  & 23.94  & +23.94 \\
\cmidrule(lr){2-6}
& \multirow{4}{*}{OOD}
& Scenario ACC     & 84.76  & 43.05  & +41.71 \\
&& Causality ACC   & 62.25  & 21.13  & +41.12 \\
&& Forecasting MAE & 186.12 & 264.18 & +78.06 \\
&& Decision ACC    & 58.97  & 23.81  & +35.16 \\

\midrule

\multirow{6}{*}{\chatts{} / Qwen2.5-14B}
& \multirow{4}{*}{A}
& Categorical score & 88.95 & 47.40 & +41.55 \\
&& Numerical score   & 78.47 & 23.53 & +54.94 \\
&& Deductive ACC     & 74.42 & 60.47 & +13.95 \\
&& Causal ACC        & 77.17 & 42.39 & +34.78 \\
\cmidrule(lr){2-6}
& \multirow{2}{*}{B}
& Categorical score  & 85.47 & 35.14 & +50.33 \\
&& Numerical score    & 78.25 & 19.38 & +58.87 \\

\midrule

\multirow{4}{*}{\tovl{} / BAGEL-7B}
& \multirow{4}{*}{OOD}
& Scenario ACC     & 81.76  & 42.83  & +38.93 \\
&& Causality ACC   & 60.63  & 28.75  & +31.88 \\
&& Forecasting MAE & 189.63 & 335.83 & +146.20 \\
&& Decision ACC    & 59.71  & 21.98  & +37.73 \\

\bottomrule
\end{tabular}
\end{table*}

\section{\benchmark{} Composition, sources, and Tasks}
\label{app:benchmark}

\subsection{Composition and sources}
\label{app:sources}

\begin{figure}[t]
\centering\includegraphics[width=0.9\linewidth]{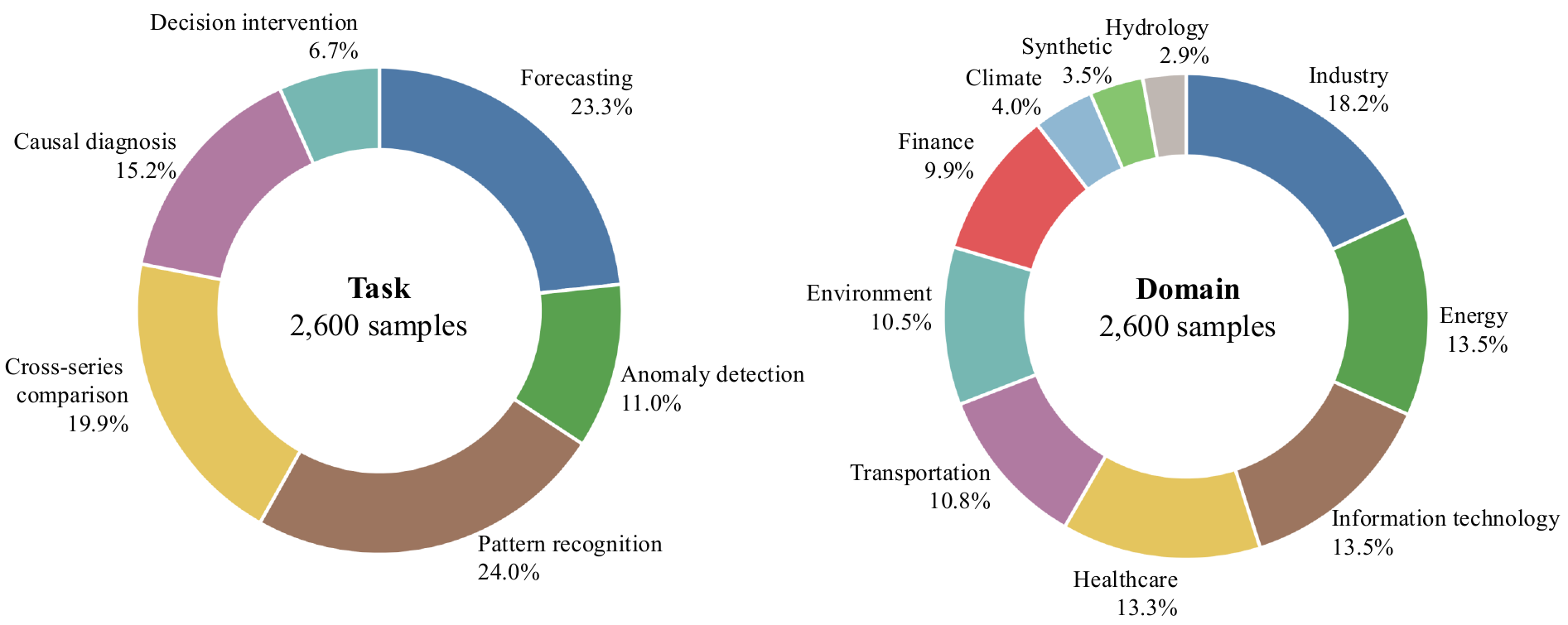}
\caption{{Tasks and domains represented in \benchmark{}.}}
\label{fig:benchmark-composition}
\end{figure}

We retain the source-defined tasks and targets, only standardize the fields needed for a common benchmark representation.
The released sources include GIFT-Eval~\citep{aksu2025gifteval}, SciTS~\citep{wu2026scits}, TIME~\citep{qiao2026time}, TSR-Suite OOD~\citep{guan2026timeomni1}, 
TSB-AD-U~\citep{liu2024elephant}, TSAQA~\citep{jing2026tsaqa}, TSRBench~\citep{yu2026tsrbench}, CaTSBench~\citep{zhou2026catsbench}, FactoryBench~\citep{factorybench2026}, and SenTSR-Bench~\citep{he2026sentsr}.
Each sample is adapted to the interface of the evaluated system, and its output is transformed to the benchmark format and evaluated using the same specific criteria as the tasks. Table~\ref{tab:benchmark-composition} contains the information about the benchmark, 
Figure~\ref{fig:benchmark-composition} summarizes the benchmark composition,
and Table~\ref{tab:source-links} lists the publicly available links.

\begin{table*}[t]
\centering
\scriptsize
\caption{Tasks, source datasets, and domains represented in
\benchmark{} after removing all Open-Meteo weather, CAMS air-quality,
and ECB exchange-rate samples.}
\label{tab:benchmark-composition}
\setlength{\tabcolsep}{3.2pt}
\renewcommand{\arraystretch}{1.16}
\resizebox{\textwidth}{!}{%
\begin{tabular}{
p{2.2cm}
p{4.6cm}
c
p{5.4cm}
}
\toprule
Task
&
Data sources
&
Samples $n$
&
Domains
\\
\midrule

Forecasting
&
GIFT-Eval;
SciTS;
TSR-Suite (OOD);
TIME
&
605
&
Energy;
environment;
finance;
healthcare;
information technology;
transportation;
climate
\\
\addlinespace


Anomaly detection
&
TSB-AD-U
&
285
&
Environment;
healthcare;
industry;
information technology;
synthetic
\\
\addlinespace

Pattern recognition
&
TSAQA;
TSRBench
&
623
&
Energy;
environment;
finance;
healthcare;
information technology;
synthetic;
transportation
\\
\addlinespace

Cross-series comparison
&
TSAQA;
CaTS-Bench
&
517
&
Energy;
environment;
finance;
healthcare;
information technology;
transportation;
climate
\\
\addlinespace

Causal diagnosis
&
FactoryBench;
SenTSR-Bench;
TSR-Suite (OOD)
&
395
&
Industry;
hydrology
\\
\addlinespace

Decision intervention
&
SenTSR-Bench;
TSR-Suite (OOD)
&
175
&
Industry;
energy
\\
\midrule

Total
&
&
2,600
&
\\
\bottomrule
\end{tabular}
}
\end{table*}

\begin{table}[t]
\centering
\scriptsize
\caption{Data sources and access links.}
\label{tab:source-links}
\setlength{\tabcolsep}{10pt}
\resizebox{\textwidth}{!}{%
\begin{tabular}{ll}
\toprule
Source & Access link \\
\midrule

GIFT-Eval &
\url{https://huggingface.co/datasets/Salesforce/GiftEval} \\

TIME &
\url{https://huggingface.co/datasets/Real-TSF/TIME} \\

SciTS &
\url{https://huggingface.co/datasets/OpenTSLab/SciTS} \\


TSB-AD-U &
\url{https://www.thedatum.org/datasets/TSB-AD-U.zip} \\

TSAQA &
\url{https://huggingface.co/datasets/TSAQA/TSAQA-Benchmark} \\

TSRBench &
\url{https://huggingface.co/datasets/umd-zhou-lab/TSRBench} \\

CaTS-Bench &
\url{https://huggingface.co/datasets/mhfisher/CaTSBench} \\

TSR-Suite &
\url{https://huggingface.co/datasets/anton-hugging/timeomni-1-testbed} \\

FactoryBench &
\url{https://huggingface.co/datasets/FactoryBench/FactoryBench} \\

SenTSR-Bench &
\url{https://huggingface.co/datasets/ZLHe0/SenTSR-Bench} \\

\bottomrule
\end{tabular}
}
\end{table}


\subsection{Tasks}
\label{app:tasks}

\paragraph{Forecasting.}
Given an observed time-series history and context provided by the source task, the goal is to predict the future sequence within the specified horizon.


\paragraph{Anomaly detection.}
For the given time series, it is required to determine whether there is any anomaly in it. 

\paragraph{Pattern recognition.}
This task is to find out the temporal pattern specified in the source sample, which may be trend, periodicity, volatility, or anomaly.

\paragraph{Cross-series comparison.}
Based on the two or more time series, the task is to identify the relationship specified in the source sample between them.

\paragraph{Causal diagnosis.}
This task is to determine the causal relationship or mechanism between the observed series and context provided by the source task.

\paragraph{Decision intervention.}
Given the observed time series and context provided by the source task, the task is to select the prescribed action or intervention.

\begin{figure}[t]
\centering
\includegraphics[width=1.0\linewidth]{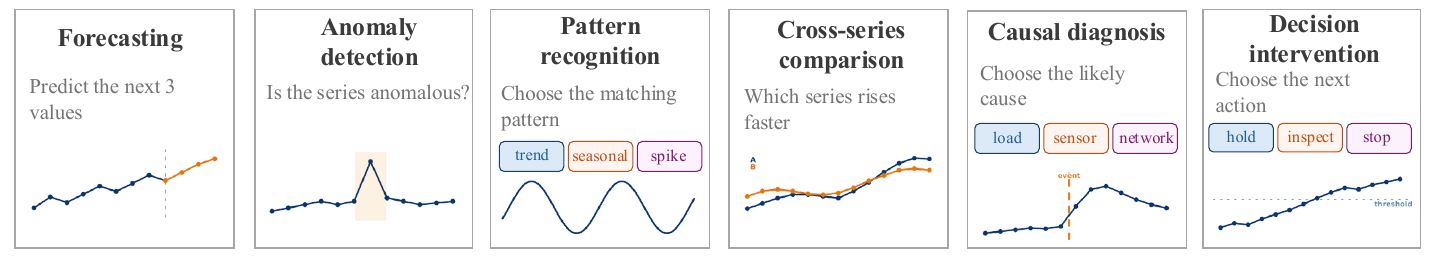}
\caption{The six tasks in \benchmark{}.}
\label{fig:tasks}
\end{figure}

Figure~\ref{fig:tasks} visualizes these six tasks. 

\subsection{Source selection and conversion}
\label{app:admission}

A source is included only if it contains samples that map to at least one of the six tasks without changing the target of the source. In order to be eligible, a source has to supply information about provenance and split in addition to either a known or independent verification of the target. An output format and scorer have to be defined for each task. All samples must satisfy all seven evaluation criteria and be rendered for all four systems without any loss of task-relevant information.

We check candidate samples from each official source split against the training data available for the evaluated systems. Overlaps are identified and removed using source IDs, fingerprints of normalized records, and grouping information. The grouping checks account for cases where the same series, episode, device, or location appears in a different form.

Before rendering inputs for specific systems, we standardize input representations, option encodings, and output
schemas while preserving the source-defined task semantics and
ground-truth answers.





\section{Intervention Semantics}
\label{app:operators}

\paragraph{Shuffle.}
Shuffle independently permutes the mutable observed values within each series. It
preserves the series length and value multiset, but destroys the original temporal
order. 

\paragraph{Flat.}
Flat replaces all mutable observed values in a series with their mean. The
replacement follows the same numerical formatting used by the benchmark, so the
displayed constant may be a rounded version of the underlying mean. Flat 
removes within-series variation while approximately preserving the observed level.

\paragraph{Noise.}
Noise replaces the mutable observed values with independent Gaussian draws using
the observed mean and population standard deviation of the series. The realized
samples are not recentered or rescaled, and therefore need not reproduce the
original sample moments exactly. When the observed standard deviation is zero, we
use a standard deviation of 1.0. Generated values are not clipped and are rendered
using the recipient's numerical precision.

\paragraph{Swap.}
Swap replaces the numerical series with a compatible series from another benchmark
record. Donors are selected without using the recipient target or model output and
are matched by source dataset, split, task contract, domain, compatibility stratum,
and number of series. A donor must come from a different record and evidence group.

\paragraph{Answer-only.}
Answer-only removes the numerical series while retaining the question and other
non-series information. Text interfaces replace each series with
\texttt{[series omitted]}, whereas native numerical interfaces receive no series
input. Question semantics, answer choices, units, length descriptions, and
mask/index definitions are left unchanged. 

\paragraph{Zero.}
Zero replaces every mutable numerical value with 0 while preserving the series
length, channel structure, and non-series information. Unlike Flat, it preserves
neither the original level nor within-series variation.

\section{Additional analysis for RQ2}
\label{app:rq2}

\subsection{Original--Answer-only prediction transitions}
\label{app:rq2-transitions}

For each task and system, we compare the semantic answers under Original and Answer-only. Correct-to-wrong and wrong-to-correct transitions are determined against the original ground-truth answer. Changed counts any difference between the two answers, including transitions between distinct incorrect answers. All three rates use the jointly valid samples as their denominator. Table~\ref{tab:rq2-all-transitions} extends the main-text analysis to all five discrete tasks. Substantial answer changes occur even in some settings with similar aggregate accuracy. These transitions identify sensitivity to evidence removal, not whether the changed answers are appropriate.

\begin{table*}[t]
\centering
\scriptsize
\caption{Original--Answer-only transitions across the five discrete tasks. C$\rightarrow$W, W$\rightarrow$C, and Changed use the joint-valid subset as denominator; all entries are percentages.}
\label{tab:rq2-all-transitions}
\setlength{\tabcolsep}{4.0pt}
\renewcommand{\arraystretch}{1.05}
\begin{tabular}{llccccc}
\toprule
Task & System & O Acc. & AO Acc. & C$\rightarrow$W & W$\rightarrow$C & Changed \\
\midrule
\multirow{4}{*}{Anomaly detection}
& TimeOmni-1  & 60.70 & 45.61 & 20.64 & 4.98 & 25.62 \\
& ChatTS      & 62.46 & 54.39 & 8.42  & 0.35 & 8.77 \\
& Time-MQA    & 64.91 & 45.61 & 20.73 & 0.36 & 21.09 \\
& TimeOmni-VL & 70.88 & 45.61 & 38.64 & 9.85 & 48.48 \\
\midrule
\multirow{4}{*}{Pattern recognition}
& TimeOmni-1  & 60.51 & 52.81 & 24.11 & 16.34 & 48.22 \\
& ChatTS      & 59.91 & 59.23 & 14.13 & 12.84 & 30.50 \\
& Time-MQA    & 55.06 & 43.98 & 17.62 & 11.88 & 34.65 \\
& TimeOmni-VL & 58.35 & 58.91 & 19.69 & 20.38 & 47.95 \\
\midrule
\multirow{4}{*}{Cross-series comparison}
& TimeOmni-1  & 58.79 & 50.87 & 20.39 & 11.37 & 33.33 \\
& ChatTS      & 54.93 & 46.23 & 15.95 & 7.85  & 25.57 \\
& Time-MQA    & 46.81 & 34.82 & 23.51 & 13.81 & 43.66 \\
& TimeOmni-VL & 53.00 & 35.59 & 41.81 & 9.70  & 55.85 \\
\midrule
\multirow{4}{*}{Causal diagnosis}
& TimeOmni-1  & 33.67 & 27.85 & 20.59 & 14.44 & 60.43 \\
& ChatTS      & 28.35 & 31.39 & 7.85  & 10.89 & 29.62 \\
& Time-MQA    & 27.19 & 19.24 & 11.78 & 11.78 & 42.42 \\
& TimeOmni-VL & 33.42 & 29.62 & 22.06 & 16.33 & 62.18 \\
\midrule
\multirow{4}{*}{Decision intervention}
& TimeOmni-1  & 43.43 & 21.14 & 32.94 & 10.59 & 62.94 \\
& ChatTS      & 26.14 & 26.29 & 5.14  & 6.29  & 25.71 \\
& Time-MQA    & 26.17 & 22.29 & 10.00 & 9.38  & 43.12 \\
& TimeOmni-VL & 49.71 & 27.43 & 33.73 & 11.45 & 66.27 \\
\bottomrule
\end{tabular}
\end{table*}




\begin{figure}[h]
\centering
\includegraphics[width=0.75\linewidth]{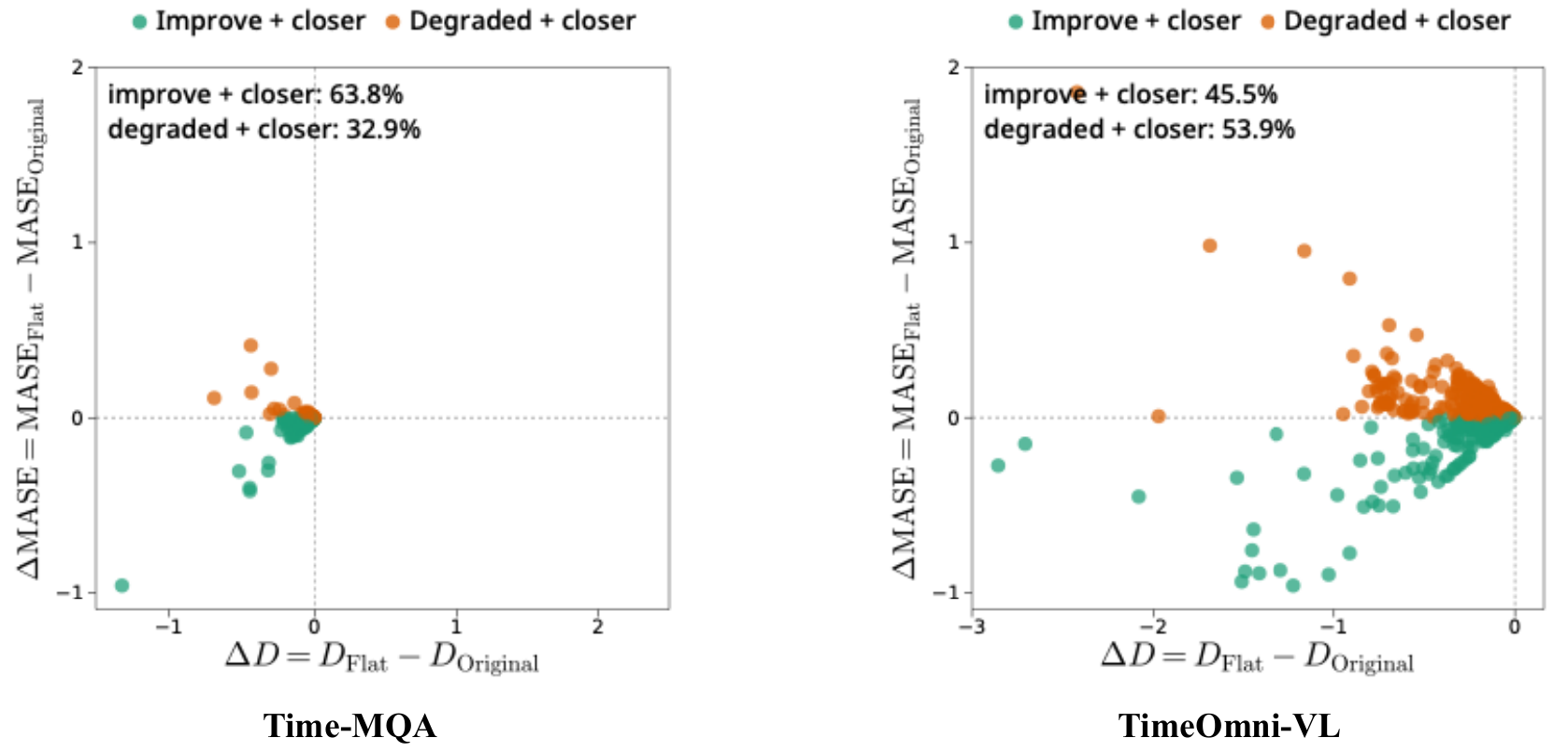}
\caption{Original--Flat item-level forecasting diagnostics for \timemqa{} and \tovl{} with available forecasting results. Panels are enlarged relative to the main-text diagnostic for readability. Negative horizontal values indicate movement toward the fixed history-mean forecast; negative vertical values indicate lower MASE under Flat.}
\label{fig:rq2-flat-all}
\end{figure}

\subsection{Flat and simple forecasting references}
\label{app:rq2-flat}

We compare Original and Flat on samples that are scorable under both conditions. Let $\bar{x}_i$ be the mean of sample $i$'s Original observed history, $H_i$ its forecast horizon, and $s_i$ the Original-history scaling term in MASE. The distance to the fixed history-mean forecast is
\[
D_i^{(c)}
=
\frac{H_i^{-1}\sum_{h=1}^{H_i}
\left|\hat{y}_{ih}^{(c)}-\bar{x}_i\right|}{s_i}.
\]
Both the reference and the scale remain fixed across conditions. We define $\Delta D_i=D_i^{\mathrm{Flat}}-D_i^{\mathrm{Original}}$ and $\Delta\mathrm{MASE}_i=\mathrm{MASE}_i^{\mathrm{Flat}}-\mathrm{MASE}_i^{\mathrm{Original}}$. Improve + closer requires both differences to be negative; Degraded + closer requires $\Delta D_i<0$ and $\Delta\mathrm{MASE}_i>0$.

Figure~\ref{fig:rq2-flat-all} shows that movement toward the history-mean forecast is common across systems, while its effect on error differs. 

\subsection{Target-aware evaluation of cross-series interventions}
\label{app:rq2-targetaware}

We recompute answers after Shuffle, Flat, Noise, Swap, and Zero for cross-series questions whose targets can be determined exactly from the transformed input. 

We distinguish pairs whose correct target changes from those whose target remains unchanged. For target-changed pairs, a correct update matches the recomputed target, a stale answer matches the original ground-truth target, Other contains all remaining outputs. These categories describe the intervention output relative to the two targets; they do not require the Original prediction to have been correct.

Table~\ref{tab:rq2-targetaware-pooled} reports accuracies over all eligible pairs and outcome shares over all target-changed pairs. Figure~\ref{fig:rq2-diagnostics}(b) uses a narrower display, normalizing correct updates and stale answers by their combined count. Table~\ref{tab:rq2-targetaware-full} separates interventions. This decomposition distinguishes incorrect responses to the transformed series from correct responses that no longer match the original answer.

\begin{table*}[t]
\centering
\small
\caption{Pooled target-aware results on the exact-oracle subset. Original-target and recomputed accuracies use all eligible pairs; the remaining columns use target-changed pairs as denominator. All entries are percentages.}
\label{tab:rq2-targetaware-pooled}
\setlength{\tabcolsep}{5pt}
\renewcommand{\arraystretch}{1.05}
\begin{tabular}{lccccc}
\toprule
System & Original-target Acc. & Recomputed Acc. & Correct update & Stale & Other \\
\midrule
TimeOmni-1  & 41.55 & 72.80 & 78.90 & 14.98 & 6.11 \\
ChatTS      & 30.49 & 68.01 & 82.87 & 6.12  & 11.01 \\
Time-MQA    & 32.88 & 48.88 & 50.76 & 18.04 & 31.19 \\
TimeOmni-VL & 39.01 & 74.44 & 77.98 & 5.50  & 16.51 \\
\bottomrule
\end{tabular}
\end{table*}

\begin{table*}[h]
\centering
\scriptsize
\caption{Target-aware decomposition by system and intervention. Target changed is the share of eligible pairs whose recomputed target changes. Correct-update, stale-answer, and other-error percentages use all target-changed outputs.}
\label{tab:rq2-targetaware-full}
\setlength{\tabcolsep}{5pt}
\renewcommand{\arraystretch}{1.05}

\begin{tabular}{llccccc}
\toprule
System
& Intervention
& Target changed
& \begin{tabular}[c]{@{}c@{}}Unchanged-target\\Acc. \end{tabular}
& Correct update
& Stale
& Other \\
\midrule

\multirow{5}{*}{TimeOmni-1}
& Shuffle &  6.7 & 62.40 & 88.89 &  0.00 & 11.11 \\
& Flat    & 76.1 & 90.62 & 94.12 &  4.90 &  0.98 \\
& Noise   & 13.4 & 73.28 & 61.11 & 27.78 & 11.11 \\
& Swap    & 50.4 & 53.03 & 62.69 & 23.88 & 13.43 \\
& Zero    & 97.8 & 66.67 & 77.10 & 17.56 &  5.34 \\
\midrule

\multirow{5}{*}{ChatTS}
& Shuffle &  6.7 & 51.20 & 44.44 & 11.11 & 44.45 \\
& Flat    & 76.1 & 84.38 & 91.18 &  1.96 &  6.86 \\
& Noise   & 13.4 & 50.86 & 33.33 & 27.78 & 38.89 \\
& Swap    & 50.4 & 50.00 & 67.16 & 16.42 & 16.42 \\
& Zero    & 97.8 & 33.33 & 93.89 &  0.76 &  5.35 \\
\midrule

\multirow{5}{*}{Time-MQA}
& Shuffle &  6.7 & 47.20 & 22.22 & 55.56 & 22.22 \\
& Flat    & 76.1 & 50.00 & 43.14 & 15.69 & 41.17 \\
& Noise   & 13.4 & 50.86 & 44.44 & 27.78 & 27.78 \\
& Swap    & 50.4 & 39.39 & 44.78 & 26.87 & 28.35 \\
& Zero    & 97.8 & 33.33 & 62.60 & 11.45 & 25.95 \\
\midrule

\multirow{5}{*}{TimeOmni-VL}
& Shuffle &  6.7 & 71.20 & 44.44 & 11.11 & 44.45 \\
& Flat    & 76.1 & 71.88 & 85.29 &  0.00 & 14.71 \\
& Noise   & 13.4 & 75.86 & 33.33 & 22.22 & 44.45 \\
& Swap    & 50.4 & 62.12 & 62.69 & 17.91 & 19.40 \\
& Zero    & 97.8 & 66.67 & 88.55 &  0.76 & 10.69 \\
\bottomrule
\end{tabular}
\end{table*}

\section{Cases}
\label{app:cases}

\input{cases/selected_case_style}


\def\casecode{1}\begin{rqcasebox}{case:selected-original-1}{Case 1. Cross-series comparison (\tomni{})}
\textbf{Question.} Given two time series A and B, detect which one has a higher amplitude defined as maximum - minimum. \par
{\small\textbf{Response labels.} \textit{a\_gt\_b}: A has the larger amplitude; \textit{a\_lt\_b}: B has the larger amplitude; \textit{tie}: equal amplitudes.\par}
\includegraphics[width=0.9\linewidth]{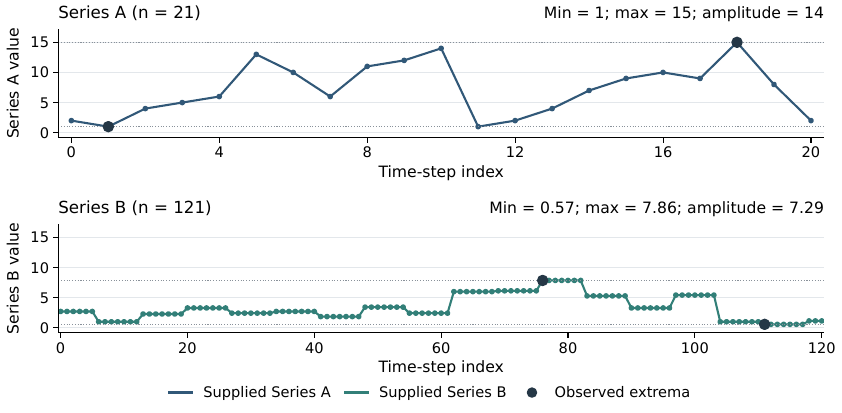}\par
{\small\textbf{Target:} \textit{a\_gt\_b} \par}
\begin{minipage}[t]{\linewidth}\small\raggedright\textbf{Answer:} \textit{a\_lt\_b} \end{minipage}\par
\begin{minipage}[t]{\linewidth}\small\raggedright\textbf{Reasoning:}\par\vspace{2pt} \vspace{4pt}
``Series A: Minimum = 1, Maximum = 15 → Amplitude = 14.''\par

``Series B: Minimum = 0.57 (last few points), Maximum = 7.86 (middle section). But the main plateau is 6.00–7.86, so the amplitude
is $7.86 - 0.57 \approx 7.29$.''\par

``Therefore, Series B has a higher amplitude than Series A.''\par
\end{minipage}\par
\vspace{3pt}\hrule\vspace{4pt}\textbf{Conflict:} Series A has amplitude 14, whereas Series B has amplitude approximately 7.29, so Series A has the larger amplitude. The rationale reverses this comparison and concludes that Series B has the higher amplitude. \par
\end{rqcasebox}

\clearpage
\def\casecode{2}\begin{rqcasebox}{case:selected-original-2}{Case 2. Pattern recognition (\tomni{})}
\textbf{Question.} Considering the time series` cyclicity, what is the approximate amplitude of its recurring peaks? \par
{\small\textbf{Options.} A: About 3.0. B: Less than 1.0. C: About 1.5. D: More than 4.0.\par}
\includegraphics[width=0.9\linewidth]{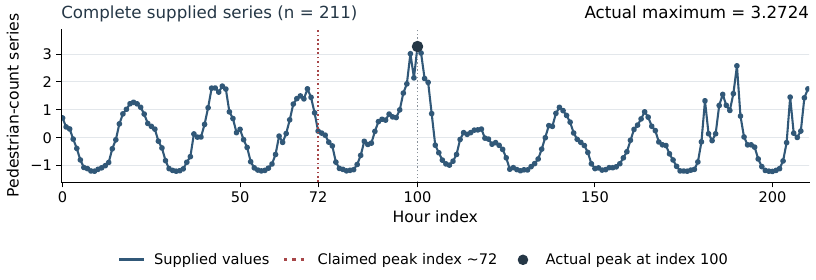}\par
{\small\textbf{Target:} A\par}
\begin{minipage}[t]{\linewidth}\small\raggedright\textbf{Answer:} B\end{minipage}\par
\begin{minipage}[t]{\linewidth}\small\raggedright\textbf{Reasoning:}\par\vspace{2pt} \vspace{4pt}
``The peak is at 3.2724 (index 72).''\par

``Option D is possible but less likely, as the highest value is 3.2724, not exceeding 4.0.''\par

``Therefore, the best estimate is Option A.''\par
\end{minipage}\par
\vspace{3pt}\hrule\vspace{4pt}\textbf{Conflict:} The rationale does not actually establish the amplitude of the recurring
peaks. It identifies a single maximum value, 3.2724, and treats that value
as the answer without examining whether multiple recurring peaks have a
similar magnitude. It also places this maximum at index 72, whereas the
value at index 72 is 0.2344 and the actual maximum occurs at index 100.
Finally, the rationale concludes that Option A is the best estimate, while
the reported final answer is B. Thus, the rationale contains both an
incorrect localization claim and an unsupported shortcut from a single
maximum to the recurring-peak amplitude, and its stated conclusion is
inconsistent with the final answer. \par
\end{rqcasebox}

\clearpage
\section{LLM-AS-JUDGE PROMPT}
\label{app:judge}

\paragraph{Human validation of the LLM-based adjudication.}
We validate the LLM-based adjudication on a frozen blinded subset
containing 240 numerical claims and 100 rationale outputs, covering both
systems and all six tasks. The human annotator does not observe the
LLM-generated labels, target answer, answer correctness, or system identity.
As shown in Table~\ref{tab:judge-human-validation}, the human annotations
show strong agreement with the LLM-based adjudication.


\begin{table}[t]
\centering
\small
\caption{
Blinded human validation of the LLM judge.
Agreement compares the LLM-generated labels with independent human annotations on a frozen validation subset covering both systems and all six tasks.
}
\label{tab:judge-human-validation}
\setlength{\tabcolsep}{5pt}
\renewcommand{\arraystretch}{1.05}
\begin{tabular}{llccc}
\toprule
Level & Criterion & $N$ & Agreement (\%) & Cohen's $\kappa$ \\
\midrule
Claim
& Factual grounding
& 240 & 95.8 & 0.920 \\

Output
& Factual grounding
& 100 & 98.0 & 0.969 \\

Output
& Inference validity
& 100 & 89.0 & 0.837 \\

Output
& Conclusion consistency
& 100 & 94.0 & 0.895 \\
\bottomrule
\end{tabular}
\end{table}

\begin{promptbox}[Prompt for factual-grounding adjudication]

You are evaluating the factual grounding of a model-generated rationale
for a time-series task.

Extract atomic claims about the observed time series that can be checked
against the supplied numerical input. Split sentences containing multiple
independently verifiable claims. Merge semantic duplicates and record their
occurrence count. Do not extract final answers, option restatements,
predicted or imputed values, recommended actions, generic method
descriptions, or external claims that cannot be verified from the supplied
series. Do not repeat claims that have already been adjudicated separately.

For each extracted claim, assign one label:

\begin{itemize}
    \item \texttt{supported}: clearly supported by the supplied numerical evidence;
    \item \texttt{unsupported}: clearly contradicted by the supplied numerical evidence;
    \item \texttt{not verifiable}: cannot be evaluated from the supplied numerical evidence.
\end{itemize}

Use only the numerical evidence provided for the current example. Do not
use the system identity, target answer, answer correctness, or answer
choices as evidence. Derived diagnostics such as trend, periodicity,
autocorrelation, anomaly, or lag may support a judgment, but should not be
treated as categorical ground truth when the claim is ambiguous.

For supported and unsupported claims, provide concise numerical evidence.
Return each unique claim once with its evidence scope, grounding label,
evidence, and occurrence count. Also indicate whether the rationale is
complete.

Return JSON matching the provided schema exactly.

\end{promptbox}

\clearpage

\section{Training-Design Diagnostics}
\label{app:training-diagnostics}

\subsection{Public training signals and release scope}

We examine official papers, supplements, repositories, model cards, and model/data releases. Table~\ref{tab:training-release} summarizes the training signals and checkpoint constraints relevant to the discussion. These observations support descriptive comparisons with the behavioral results.

None of the public releases provides the complete sequence of intermediate training checkpoints. Time-MQA permits a fixed-base contrast because its specialization is released as a separate LoRA adapter, although the original base revision is not pinned. We distinguish this comparison from an exact training-stage ablation.

\begin{table}[t]
\centering
\small
\caption{Public training signals and checkpoint constraints relevant to the discussion. Missing public artifacts limit the comparisons supported by the released systems.}
\label{tab:training-release}
\setlength{\tabcolsep}{4pt}
\renewcommand{\arraystretch}{1.08}
\begin{tabular}{@{}p{0.14\linewidth}p{0.44\linewidth}p{0.36\linewidth}@{}}
\toprule
System & Public training signals & Checkpoint constraints \\
\midrule
TimeOmni-1 & Human-guided CoT-SFT and GRPO, with answer/format rewards and a forecasting-error reward. & Final model available; no released SFT-only or RL-intermediate checkpoint. \\
ChatTS & Alignment and supervised fine-tuning; the training data released with the paper include multi-series, relational, causal, and rationale-like content. & Historical training data and final model available; no released alignment-only checkpoint. \\
Time-MQA & Reported training uses 7,000 TSQA and 3,000 OpenOrca samples; final specialization is a LoRA adapter. & Adapter permits a fixed-base contrast; the original base revision is not pinned and no intermediate adapter is released. \\
TimeOmni-VL & Time-series understanding/generation and CoT-based post-training. & Final model available; no intermediate checkpoints or complete original training-row manifest. \\
\bottomrule
\end{tabular}
\end{table}

\subsection{ChatTS training corpus}
\label{app:chatts-corpus}

We scan the historical Stage~1 and Stage~2 files, containing 105,085 and 30,650 samples, respectively. Overlapping rule-based tags identify candidate multi-series, cross-series, causal, and rationale-like content. We manually check 57 stage-by-positive-category samples to assess the semantics of the tagging rules. Table~\ref{tab:chatts-corpus} reports the categories used in the discussion; the tags are not gold annotations.

The public generation pipeline includes numerical statements and source-series attributes in training text, and TSEvol applies holistic LLM filtering to complete QA pairs. This does not establish that every intermediate numerical statement is checked: our audit identifies no general deterministic per-claim numerical checker applying throughout the pipeline described in the paper.

We also inspect 50 candidate groups, split evenly across stages, for matched questions with changed numerical evidence and recomputed targets. None is verified as a same-parent evidence-change/recomputed-target training pair. This finding is limited to the audited candidates and does not establish that such supervision was absent from private training procedures.

\begin{table}[t]
\centering
\small
\caption{Selected candidate tags in the ChatTS corpus. Samples gives the stage size; all other entries are percentages within that stage. Rule-based tags overlap and are not gold annotations.}
\label{tab:chatts-corpus}
\setlength{\tabcolsep}{4pt}
\begin{tabular}{lccccc}
\toprule
Stage & Samples & Multi-series & Cross-series & Causal & Rationale \\
\midrule
Stage 1 & 105,085 & 66.69 & 6.89 & 2.27 & 51.15 \\
Stage 2 & 30,650 & 29.34 & 19.61 & 10.11 & 42.39 \\
\bottomrule
\end{tabular}
\end{table}

\subsection{Time-MQA Base--LoRA checkpoint contrast}
\label{app:timemqa-contrast}

\paragraph{Setup.}
We freeze one public revision of the base declared by the released adapter and compare it with the same base plus the LoRA weights. The tokenizer, model-visible input, prompt renderer, greedy decoding, parser, and scorer are held identical. Invalid outputs remain in the overall accuracy denominator. Because the adapter configuration does not identify the exact original base revision, this is a reproducible checkpoint contrast rather than a reconstruction of the authors' training stages. Forecasting is excluded from the performance interpretation because only two samples are scorable under both checkpoints.

\paragraph{Target-aware responses.}
Both checkpoints are evaluated on the same 669 eligible cross-series sample--intervention pairs, of which 327 have a changed correct target. Table~\ref{tab:timemqa-targetaware-pooled} reports original-target and recomputed-target accuracies over all 669 pairs, and the three outcome shares over the 327 target-changed pairs. Table~\ref{tab:timemqa-targetaware-intervention} provides the intervention-level results. Adaptation increases correct updates and reduces stale answers in the pooled comparison, with the largest correct-update increases under Flat and Zero rather than a uniform improvement across interventions.

\paragraph{Validity and paired-valid accuracy.}
Output validity is the proportion of task samples yielding recoverable semantic answers. Paired-valid accuracy is evaluated on the common subset for which both checkpoints produce valid answers. Table~\ref{tab:timemqa-validity} reports these measures alongside overall accuracy. Pattern recognition shows higher validity but lower paired-valid accuracy after adaptation, whereas causal diagnosis improves on both measures. Decision intervention increases validity without changing paired-valid accuracy, and cross-series comparison changes little. The task-specific patterns show why aggregate accuracy gains cannot be interpreted as a uniform change in semantic correctness.

\begin{table}[t]
\centering
\small
\caption{
Pooled Time-MQA target-aware Base--LoRA contrast. Original-target and recomputed-target accuracies use all 669 eligible sample--intervention pairs. Correct update, stale, and other use the 327 pairs whose correct target changes. All entries are percentages.
}
\label{tab:timemqa-targetaware-pooled}
\setlength{\tabcolsep}{4.5pt}
\renewcommand{\arraystretch}{1.08}
\begin{tabular}{lrrrrr}
\toprule
Checkpoint
& Original-target Acc.
& Recomputed Acc.
& Correct update
& Stale
& Other \\
\midrule
Base & 39.91 & 27.06 & 17.13 & 43.43 & 39.45 \\
LoRA & 24.81 & 38.27 & 44.04 & 16.51 & 39.45 \\
\bottomrule
\end{tabular}
\end{table}

\begin{table*}[t]
\centering
\small
\caption{
Intervention-level \timemqa{} target-aware behavior. ``Changed'' is the number
of eligible pairs for which the intervention changes the recomputed target.
Correct-update, stale, and other percentages use the changed-target subset.
}
\label{tab:timemqa-targetaware-intervention}
\setlength{\tabcolsep}{4pt}
\renewcommand{\arraystretch}{1.05}

\resizebox{\textwidth}{!}{%
\begin{tabular}{llcccccccc}
\toprule
Checkpoint
& Intervention
& Eligible
& Changed
& Original-target Acc.
& Recomputed Acc.
& Correct update
& Stale
& Other \\
\midrule

\multirow{5}{*}{Base}
& Shuffle & 134 &   9 & 38.06 & 35.07 & 11.11 & 55.56 & 33.33 \\
& Flat    & 134 & 102 & 36.57 &  8.96 &  1.96 & 38.24 & 59.80 \\
& Noise   & 134 &  18 & 37.31 & 40.30 & 33.33 & 11.11 & 55.56 \\
& Swap    & 133 &  67 & 40.60 & 30.08 & 28.36 & 49.25 & 22.39 \\
& Zero    & 134 & 131 & 47.01 & 20.90 & 21.37 & 48.09 & 30.54 \\
\midrule

\multirow{5}{*}{LoRA}
& Shuffle & 134 &   9 & 32.09 & 29.85 & 22.22 & 55.56 & 22.22 \\
& Flat    & 134 & 102 & 19.40 & 41.04 & 41.18 & 12.75 & 46.07 \\
& Noise   & 134 &  18 & 35.07 & 34.33 & 16.67 & 22.22 & 61.11 \\
& Swap    & 133 &  67 & 26.32 & 28.57 & 29.85 & 25.37 & 44.78 \\
& Zero    & 134 & 131 & 11.19 & 57.46 & 58.78 & 11.45 & 29.77 \\
\bottomrule
\end{tabular}%
}
\end{table*}



\begin{table*}[t]
\centering
\small
\caption{
\timemqa{} validity and paired-valid accuracy. Paired-valid accuracy uses only
samples for which both checkpoints produce valid semantic outputs.
}
\label{tab:timemqa-validity}
\setlength{\tabcolsep}{4.0pt}
\renewcommand{\arraystretch}{1.08}
\begin{tabular}{lcccccc}
\toprule
& \multicolumn{2}{c}{Valid (\%)}
& \multicolumn{2}{c}{Paired-valid acc. (\%)}
& \multicolumn{2}{c}{Overall acc. (\%)} \\
\cmidrule(lr){2-3}
\cmidrule(lr){4-5}
\cmidrule(lr){6-7}
Task
& Base & LoRA
& Base & LoRA
& Base & LoRA \\
\midrule
Anomaly detection
& 98.95 & 92.63 & 55.17 & 65.90 & 56.84 & 61.05 \\
Pattern recognition
& 26.97 & 91.49 & 73.91 & 53.42 & 20.06 & 53.45 \\
Cross-series comparison
& 26.69 & 25.73 & 50.00 & 48.68 & 13.73 & 13.35 \\
Causal diagnosis
& 50.38 & 86.08 & 23.93 & 35.58 & 11.65 & 25.06 \\
Decision intervention
& 36.00 & 90.86 & 24.56 & 24.56 & 9.71 & 21.14 \\
\bottomrule
\end{tabular}
\end{table*}

\end{document}